\documentclass[letterpaper,11pt]{article}
\usepackage[margin=1in,letterpaper]{geometry} 

\usepackage{hyperref}
\usepackage{url}

\usepackage{bm}
\usepackage{natbib}
\usepackage{subfigure}
\usepackage{amsmath}
\usepackage{amsthm}
\usepackage{amssymb}
\usepackage{amsfonts}
\usepackage{amsopn}
\usepackage{verbatim}
\usepackage{algorithmic,algorithm}
\usepackage{color}
\usepackage{enumitem}
\usepackage{stfloats}
\usepackage{float}
\usepackage{multirow}
\usepackage{siunitx}
\usepackage{graphicx}
\usepackage{mathrsfs}
\usepackage{thmtools,thm-restate}
\usepackage{threeparttable}

\usepackage{amsthm}
\usepackage{bm}
\usepackage{color}
\usepackage{enumitem}
\usepackage{stfloats}
\usepackage{float}
\usepackage{multirow}
\usepackage{siunitx}
\usepackage{amsfonts}
\usepackage{amsmath}
\usepackage{amssymb}
\usepackage{mathtools}
\usepackage{amsthm}
\usepackage{wrapfig}

\usepackage{url}
\usepackage{natbib}
\usepackage{threeparttable}
\newtheorem{theorem}{Theorem}
\newtheorem{corollary}{Corollary}[theorem]
\newtheorem{proposition}{Proposition}

\title{Revisiting Global Pooling through the Lens of Optimal Transport}

\author{Minjie Cheng$^*$~~and~~Hongteng Xu\thanks{Equal contribution} \\
Gaoling School of Artificial Intelligence\\
Renmin University of China\\
Beijing, China \\
\texttt{\{chengminjie,hongtengxu\}@ruc.edu.cn} \\
}

\begin{document}

\maketitle

\begin{abstract}
Global pooling is one of the most significant operations in many machine learning models and tasks, whose implementation, however, is often empirical in practice. 
In this study, we develop a novel and solid global pooling framework through the lens of optimal transport. 
We demonstrate that most existing global pooling methods are equivalent to solving some specializations of an unbalanced optimal transport (UOT) problem. 
Making the parameters of the UOT problem learnable, we unify various global pooling methods in the same framework, and accordingly, propose a generalized global pooling layer called UOT-Pooling (\textit{UOTP}) for neural networks. 
Besides implementing the UOTP layer based on the classic Sinkhorn-scaling algorithm, we design a new model architecture based on the Bregman ADMM algorithm, which has better numerical stability and can reproduce existing pooling layers more effectively.
We test our UOTP layers in several application scenarios, including multi-instance learning, graph classification, and image classification.
Our UOTP layers can either imitate conventional global pooling layers or learn some new pooling mechanisms leading to better performance.
\end{abstract}

\section{Introduction}
As an essential operation of information fusion, global pooling aims to achieve a global representation for a set of inputs and make the representation invariant to the permutation of the inputs. 
This operation has been widely used in many machine learning models. 
For example, we often leverage a global pooling operation to aggregate multiple instances into a bag-level representation in multi-instance learning tasks~\citep{ilse2018attention,yan2018deep}.
Another example is graph embedding. 
Graph neural networks apply various pooling layers to merge node embeddings into a global graph embedding~\citep{ying2018hierarchical,xu2018powerful}. 
Besides these two cases, global pooling is also necessary for convolutional neural networks~\citep{krizhevsky2012imagenet,he2016deep}. 
Therefore, the design of global pooling operation is a fundamental problem for many applications.

Nowadays, simple global pooling operations like mean-pooling (or called average-pooling) and max-pooling~\citep{boureau2010theoretical} are commonly used because of their computational efficiency. 
The mixture and the concatenation of these simple operations are also considered to improve their performance~\citep{lee2016generalizing}. 
Recently, many pooling methods, $e.g.$, Network-in-Network (NIN)~\citep{lin2013network}, Set2Set~\citep{vinyals2015order}, DeepSet~\citep{zaheer2017deep}, attention-pooling~\citep{ilse2018attention}, and SetTransformer~\citep{lee2019set}, are developed with learnable parameters and more sophisticated mechanisms. 
Although the above pooling methods work well in many scenarios, their theoretical study is far lagged-behind --- the principles of the methods are not well-interpreted, whose rationality and effectiveness are not supported in theory. 
Without insightful theoretical guidance, the design and the selection of global pooling are empirical and time-consuming, often leading to suboptimal performance in practice.

\begin{figure}[t]
    \centering
    \subfigure[The principle of our UOTP layer]{
    \includegraphics[height=3.75cm]{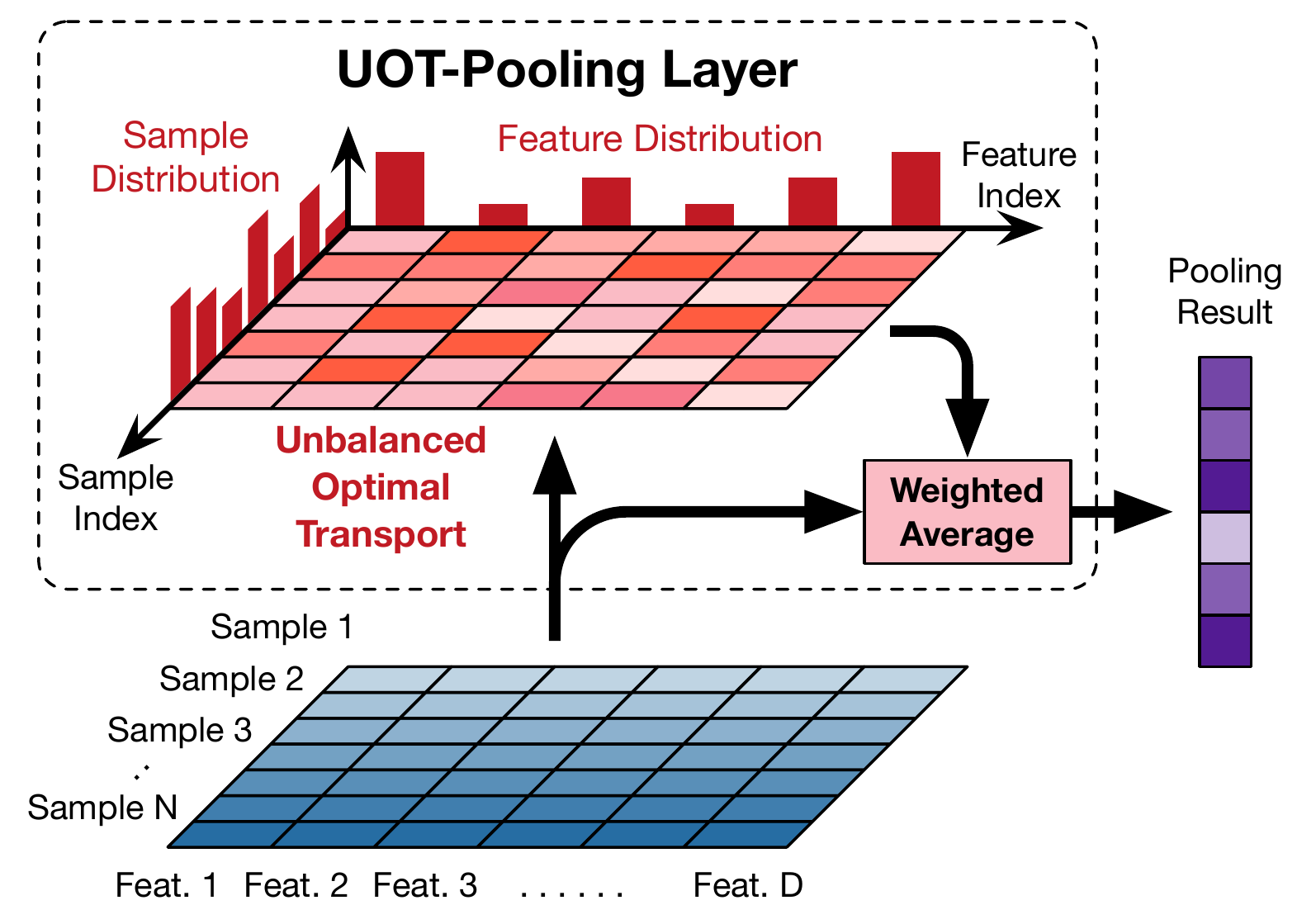}\label{fig:scheme}
    }
    \hspace{-5pt}
    \subfigure[The BADMM-based UOTP layer]{
    \includegraphics[height=3.75cm]{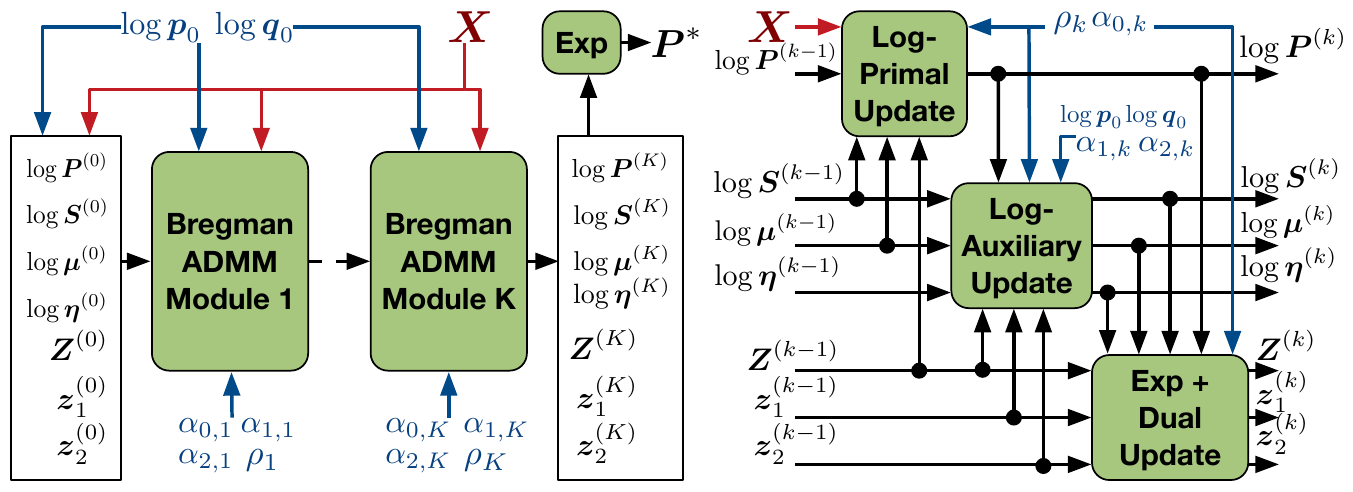}\label{fig:badmm_uot}
    }
    \vspace{-5pt}
    \caption{(a) An illustration of the proposed UOTP layer. 
    (b) The BADMM-based UOTP layer (left) and a single BADMM module (right). 
    The input, model parameters, and intermediate variables are labeled in red, blue, and black. 
    More details are shown in Section~\ref{ssec:badmm} and Appendix~\ref{app:badmm}.
    }
\end{figure}

In this study, we propose a novel algorithmic global pooling framework to unify and generalize many existing global pooling operations through the lens of optimal transport. 
As illustrated in Figure~\ref{fig:scheme}, we revisit a pooling operation from the viewpoint of optimization, formulating it as optimizing the joint distribution of sample index and feature dimension for weighting and averaging representative ``sample-feature'' pairs. 
From the viewpoint of statistical signal processing, this framework achieves global pooling based on the expectation-maximization principle.
We show that the proposed optimization problem corresponds to an unbalanced optimal transport (UOT) problem. 
Moreover, we demonstrate that most existing global pooling operations are specializations of the UOT problem under different parameter configurations.

By making the parameters of the UOT problem learnable, we design a new generalized global pooling layer for neural networks, called UOT-Pooling (or \textit{UOTP} for short).
Its forward computation corresponds to solving the UOT problem, while the backpropagation step updates the parameters of the problem. 
Besides implementing the UOTP layer based on the well-known Sinkhorn-scaling algorithm~\citep{cuturi2013sinkhorn,pham2020unbalanced}, we design a new model architecture based on the Bregman alternating direction method of multipliers (Bregman ADMM, or BADMM for short)~\citep{wang2014bregman,xu2020gromov}, as shown in Figure~\ref{fig:badmm_uot}. 
Each implementation unrolls the iterative optimization steps of the UOT problem, whose complexity and stability are analyzed quantitatively. 
In summary, the contributions of our work include three folds.

\textbf{Modeling.} To our knowledge, we make the first attempt to propose a unified global pooling framework from the viewpoint of computational optimal transport. 
The proposed UOTP layer owns the permutation-invariance property and can cover typical global pooing methods.
    
\textbf{Algorithm.} We propose a Bregman ADMM algorithm to solve the UOT problem and implement a UOTP layer based on it.
Compared to the UOTP implemented based on the Sinkhorn-scaling algorithm, our BADMM-based UOTP layer owns better numerical stability and learning performance. 

\textbf{Application.}  We test our UOTP layer in multi-instance learning, graph classification, and image classification. 
In most situations, our UOTP layers either are comparable to conventional pooling methods or outperform them, and thus simplify the design and selection of global pooling.

\section{Proposed UOT-Pooling Framework}
\subsection{A generalized formulation of global pooling operations}

Denote $\mathcal{X}_D=\{\bm{X}\in \mathbb{R}^{D\times N}|N\in\mathbb{N}\}$ as the space of sample sets. 
Each $\bm{X}=[\bm{x}_1,...,\bm{x}_N]\in\mathbb{R}^{D\times N}$ contains $N$ $D$-dimensional feature vectors. 
A global pooling operation $f:\mathcal{X}_{D}\mapsto\mathbb{R}^D$ maps each set to a single vector and ensures the output is \textit{permutation-invariant}, $i.e.$, $f(\bm{X})=f(\bm{X}_{\pi})$ for $\bm{X},\bm{X}_{\pi}\in\mathcal{X}_D$, where $\bm{X}_{\pi}=[\bm{x}_{\pi(1)},...,\bm{x}_{\pi(N)}]$ and $\pi$ is an arbitrary permutation. 
Following the work in~\citep{gulcehre2014learned,li2020deepergcn,ko2021learning}, we assume the input data $\bm{X}$ to be nonnegative. 
Note that, this assumption is reasonable in general because the input data is often processed by nonnegative activations, like ReLU, sigmoid, and so on.
For some pooling methods, $e.g.$, the max-pooling shown below, the nonnegativeness is even necessary.

Typically, the widely-used mean-pooling takes the average of the input vectors as its output, $i.e.$, $f(\bm{X})=\frac{1}{N}\sum_{n=1}^N\bm{x}_n$. 
Another popular pooling operation, max-pooling, concatenates the maximum of each dimension as its output, $i.e.$, $f(\bm{X})=\|_{d=1}^{D} \max_n\{x_{dn}\}_{n=1}^{N}$, where $x_{dn}$ is the $d$-th element of $\bm{x}_n$ and ``$\|$'' represents the concatenation operator. 
The attention-pooling in~\citep{ilse2018attention} derives a vector on the $(N-1)$-Simplex from the input $\bm{X}$ and outputs the weighted summation of the input vectors, $i.e.$, $f(\bm{X})=\bm{X}\bm{a}_{\bm{X}}$ and $\bm{a}_{\bm{X}}=\text{softmax}(\bm{w}^T\text{tanh}(\bm{VX}))^T\in\Delta^{N-1}$. 

For each $\bm{X}$, its element $x_{dn}$ corresponds to a ``sample-feature'' pair. 
Essentially, the above global pooling operations would like to predict the significance of such pairs and output their weighted column-wise average. 
In particular, denote $\bm{P}=[p_{dn}]\in [0, 1]^{D\times N}$ as the joint distribution of the sample index and the feature dimension. 
We obtain a generalized formulation of global pooling:
\begin{eqnarray}\label{eq:pooling}
f(\bm{X})=(\bm{X}\odot \underbrace{\text{diag}^{-1}(\bm{P1}_N)\bm{P}}_{\tilde{\bm{P}}=[p_{n|d}]})\bm{1}_N
=\big\Vert_{d=1}^{D}\mathbb{E}_{n\sim p_{n|d}}[x_{dn}],
\end{eqnarray}
where $\odot$ is the Hadamard product, $\text{diag}(\cdot)$ converts a vector to a diagonal matrix, and $\bm{1}_N$ represents the $N$-dimensional all-one vector. 
$\bm{P1}_N=\bm{p}$ is the marginal distribution of $\bm{P}$ corresponding to feature dimensions. 
$\text{diag}^{-1}(\bm{p})\bm{P}=\tilde{\bm{P}}=[p_{n|d}]$ normalizes the rows of $\bm{P}$, and the $d$-th row leads to the distribution of sample indexes conditioned on the $d$-th feature dimension. 
Therefore, we can interpret~(\ref{eq:pooling}) as calculating and concatenating the conditional expectation of $x_{dn}$'s for $d=1,...,D$.

Different pooling operations derive $\bm{P}$ based on different weighting mechanisms. 
Mean-pooling treats each element evenly, and thus, $\bm{P}=[\frac{1}{DN}]$. 
Max-pooling sets $\bm{P}\in \{0, \frac{1}{D}\}^{D\times N}$ and $p_{dn}=\frac{1}{D}$ if and only if $n=\arg\max_{m}\{x_{dm}\}_{m=1}^{N}$. 
Attention-pooling derives $\bm{P}$ as a learnable rank-one matrix, $i.e.$, $\bm{P}=\frac{1}{D}\bm{1}_D\bm{a}^{T}_{\bm{X}}$. 
All these operations set the marginal distribution $\bm{p}=\bm{P1}_N$ to be uniform, $i.e.$, $\bm{p}=[\frac{1}{D}]$, while let the other marginal distribution $\bm{q}=\bm{P}^T\bm{1}_D$ unconstrained.

\subsection{Global pooling via solving unbalanced optimal transport problem}

The above analysis implies that we can unify typical pooling operations in an interpretable algorithmic framework, in which all these operations aim at deriving the joint distribution $\bm{P}$. 
From the viewpoint of statistical signal processing~\citep{turin1960introduction}, the input signal $\bm{X}$ is modulated by $\bm{P}$. 
To keep the modulated signal as informative as possible, many systems, $e.g.$, antenna arrays in telecommunication systems, keep or enlarge its expected amplitude.
Following this ``expectation-maximization'' principle, we learn $\bm{P}$ to maximize the expectation in~(\ref{eq:pooling}):
\begin{eqnarray}\label{eq:ot}
\bm{P}^*=\arg\sideset{}{_{\bm{P}\in \Pi(\bm{p},\bm{q})}}\max \sideset{}{_{d=1}^{D}}\sum p_d \mathbb{E}_{n\sim p_{n|d}}[x_{dn}] =\arg\sideset{}{_{\bm{P}\in \Pi(\bm{p},\bm{q})}}\max\underbrace{\mathbb{E}_{(d, n)\sim \bm{P}}[x_{dn}]}_{\langle\bm{X},\bm{P}\rangle},
\end{eqnarray}
where $\langle\cdot,\cdot\rangle$ represents the inner product of matrices. 
$\bm{p}\in\Delta^{D-1}$ and $\bm{q}\in\Delta^{N-1}$ are the distribution of feature dimension and that of sample index, respectively, which determine the marginal distributions of $\bm{P}$, $i.e.$, $\bm{P}\in \Pi(\bm{p},\bm{q})=\{\bm{P}\geq \bm{0}|\bm{P}\bm{1}_N=\bm{p},\bm{P}^T\bm{1}_D=\bm{q}\}$. 

Through~(\ref{eq:ot}), we have connected the global pooling problem to computational optimal transport --- (\ref{eq:ot}) is an optimal transport problem~\citep{villani2008optimal}, which learns the optimal joint distribution $\bm{P}^*$ to maximize the expectation of $x_{dn}$.
Plugging $\bm{P}^*$ into~(\ref{eq:pooling}) leads to a global pooling result of $\bm{X}$. 
Note that, achieving global pooling merely based on~(\ref{eq:ot}) often suffers from some limitations in practice. 
Firstly, solving~(\ref{eq:ot}) is time-consuming and always leads to sparse solutions because it is a constrained linear programming problem. 
A sparse $\bm{P}^*$ tends to filter out some weak but possibly-informative values in $\bm{X}$, which may do harm to downstream tasks.
Secondly, solving~(\ref{eq:ot}) requires us to know the marginal distributions $\bm{p}$ and $\bm{q}$ in advance, which is either infeasible or too strict in practice. 

To make the framework feasible in practice, we improve the smoothness of $\bm{P}^*$ and introduce two prior distributions ($i.e.$, $\bm{p}_0$ and $\bm{q}_0$) to regularize the marginals of $\bm{P}^*$, which leads to the following unbalanced optimal transport (UOT) problem~\citep{benamou2015iterative,pham2020unbalanced}:
\begin{eqnarray}\label{eq:uot}
\begin{aligned}
\bm{P}_{\text{uot}}^*(\bm{X};\bm{\theta})
=\arg\sideset{}{_{\bm{P}}}\min\langle-\bm{X},\bm{P}\rangle + 
\alpha_0 \text{R}(\bm{P}) +
\alpha_1\text{KL}(\bm{P1}_N | \bm{p}_0)+\alpha_2 \text{KL}(\bm{P}^T\bm{1}_D | \bm{q}_0).
\end{aligned}
\end{eqnarray}
Here, $\text{R}(\bm{P})$ is a smoothness regularizer making the optimal transport problem strictly-convex, whose significance is controlled by $\alpha_0$. 
We often set the regularizer to be entropic~\citep{cuturi2013sinkhorn}, $i.e.$, $\text{R}(\bm{P})=\langle\bm{P},\log\bm{P}-\bm{1}\rangle=\sum_{d,n}p_{dn}(\log p_{dn} -1)$, or quadratic~\citep{blondel2018smooth}, $i.e.$, $\text{R}(\bm{P})=\langle\bm{P},\bm{P}\rangle$. 
$\text{KL}(\bm{a}|\bm{b})=\langle\bm{a},\log\bm{a}-\log\bm{b}\rangle-\langle\bm{a}-\bm{b},\bm{1}\rangle$ represents the KL-divergence between $\bm{a}$ and $\bm{b}$. 
The two KL-based regularizers in~(\ref{eq:uot}) penalize the differences between the marginals of $\bm{P}$ and the prior distributions $\bm{p}_0$ and $\bm{q}_0$, whose significance is controlled by $\alpha_1$ and $\alpha_2$, respectively. 
For convenience, we use $\bm{\theta}=\{\alpha_0,\alpha_1,\alpha_2,\bm{p}_0,\bm{q}_0\}$ to represent the model parameters.

As shown in~(\ref{eq:uot}), the optimal transport $\bm{P}_{\text{uot}}^*$ can be viewed as a function of $\bm{X}$, whose parameters are the weights of the regularizers and the prior distributions, $i.e.$, $\bm{P}_{\text{uot}}^*(\bm{X};\bm{\theta})$. 
Plugging it into~(\ref{eq:pooling}), we obtain the proposed UOT-Pooling operation:
\begin{eqnarray}\label{eq:uotp}
\begin{aligned}
f_{\text{uot}}(\bm{X};\alpha_0,\alpha_1,\alpha_2,\bm{p}_0,\bm{q}_0) 
= (\bm{X}\odot (\text{diag}^{-1}( \bm{P}_{\text{uot}}^*(\bm{X};\bm{\theta})\bm{1}_N)\bm{P}_{\text{uot}}^*(\bm{X};\bm{\theta})))\bm{1}_N,
\end{aligned}
\end{eqnarray}

Our UOT-Pooling satisfies the requirement of permutation-invariance under mild conditions. 
\begin{theorem}\label{thm:pi}
The UOT-Pooling in~(\ref{eq:uotp}) is permutation-invariant, $i.e.$, $f_{\text{uot}}(\bm{X})=f_{\text{uot}}(\bm{X}_\pi)$ for an arbitrary permutation $\pi$, when the $\bm{q}_0$ in~(\ref{eq:uot}) is a permutation-equivariant function of $\bm{X}$.
\end{theorem} 
\begin{corollary}\label{coro:pi}
The UOT-Pooling in~(\ref{eq:uotp}) is permutation-invariant when the $\bm{q}_0$ in~(\ref{eq:uot}) is uniform, $i.e.$, $\bm{q}_0=\frac{1}{N}\bm{1}_N$ for any $\bm{X}\in\mathbb{R}^{D\times N}$.
\end{corollary}

\subsection{Connecting to representative pooling operations}
Our UOT-Pooling provides a unified pooling framework. 
In particular, we demonstrate that many existing pooling operations can be formulated as the specializations of~(\ref{eq:uotp}) under different settings.
\begin{proposition}[UOT for typical pooling operations]\label{prop:equiv}
Given an arbitrary $\bm{X}\in\mathbb{R}^{D\times N}$, the mean-pooling, max-pooling, and the attention-pooling with attention weights $\bm{a}_{\bm{X}}$ can be equivalently achieved by the $f_{\text{uot}}(\bm{X};\alpha_0,\alpha_1,\alpha_2,\bm{p}_0,\bm{q}_0)$ in~(\ref{eq:uotp}) under the following configurations:
\begin{center}
\begin{small}
\begin{tabular}{@{\hspace{3pt}}l@{\hspace{3pt}}|@{\hspace{3pt}}l@{\hspace{3pt}}}
\hline
Pooling methods & $f_{\text{uot}}(\bm{X};\alpha_0,\alpha_1,\alpha_2,\bm{p}_0,\bm{q}_0)$\\
\hline
Mean-pooling & $\alpha_0,\alpha_1,\alpha_2\rightarrow\infty$, $\bm{p}_0=\frac{1}{D}\bm{1}_D$, $\bm{q}_0=\frac{1}{N}\bm{1}_N$\\
Max-pooling & $\alpha_0,\alpha_2\rightarrow 0$, $\alpha_1\rightarrow\infty$, $\bm{p}_0=\frac{1}{D}\bm{1}_D$, $\bm{q}_0=-$ \\
Attention-pooling & $\alpha_0,\alpha_1,\alpha_2\rightarrow\infty$, $\bm{p}_0=\frac{1}{D}\bm{1}_D$, $\bm{q}_0=\bm{a}_{\bm{X}}$\\
\hline
\end{tabular}
\end{small}
\end{center}
Here, ``$\bm{q}_0=-$'' means that $\bm{q}_0$ is unconstrained, and $\alpha_{1},\alpha_{2}\rightarrow\infty$ means the regularizers become strict equality constraints, rather than ignoring the optimal transport term $\langle -\bm{X},\bm{P}\rangle$. 
\end{proposition}

Additionally, the combination of such UOT-Pooling operations reproduces other pooling mechanisms, such as the mixed mean-max pooling operation in~\citep{lee2016generalizing}:
\begin{eqnarray}\label{eq:mmp}
f_{\text{mix}}(\bm{X})=\omega \text{MeanPool}(\bm{X}) + (1-\omega)\text{MaxPool}(\bm{X}).
\end{eqnarray}
When $\omega\in (0,1)$ is a learnable scalar,~(\ref{eq:mmp}) is called ``Mixed mean-max pooling". 
When $\omega$ is parameterized as a sigmoid function of $\bm{X}$,~(\ref{eq:mmp}) is called ``Gated mean-max pooling". 
Such mixed pooling operations can be achieved by integrating three UOT-Pooling operations in a hierarchical way:
\begin{proposition}[Hierarchical UOT for mixed pooling]\label{prop:mix}
Given an arbitrary $\bm{X}\in\mathbb{R}^{D\times N}$, the $f_{\text{mix}}(\bm{X})$ in~(\ref{eq:mmp}) can be equivalently implemented by $f_{\text{uot}}([f_{\text{uot}}(\bm{X};\bm{\theta}_1),f_{\text{uot}}(\bm{X};\bm{\theta}_2)];\bm{\theta}_3)$, 
where $\bm{\theta}_1=\{\infty,\infty,\infty,\frac{1}{D}\bm{1}_D,\frac{1}{N}\bm{1}_N\}$, $\bm{\theta}_2=\{0, \infty,0,\frac{1}{D}\bm{1}_D,-\}$, and $\bm{\theta}_3=\{\infty,\infty,\infty,\frac{1}{D}\bm{1}_D,[\omega,1-\omega]^T\}$.
\end{proposition}

The proofs of Theorem~\ref{thm:pi}, Corollary~\ref{coro:pi}, and above Propositions are given in Appendix~\ref{app:proof}.
 
\section{Implementing Learnable UOT-Pooling Layers}
Beyond reproducing existing pooling operations, we can implement the UOT-Pooling as a learnable neural network layer, whose feed-forward computation solves~(\ref{eq:uot}) and parameters can be learned via the backpropagation. 
Typically, when the smoothness regularizer is entropic, we can implement the UOTP layer based on the Sinkhorn scaling algorithm~\citep{chizat2018scaling,pham2020unbalanced}.
This algorithm solves the dual problem of~(\ref{eq:uot}) iteratively: 
$i$) Initialize dual variables as $\bm{a}^{(0)}=\bm{0}_D$ and $\bm{b}^{(0)}=\bm{0}_N$. 
$ii$) In the $k$-th iteration, update current dual variables $\bm{a}^{(k)}$ and $\bm{b}^{(k)}$ by
\begin{eqnarray}\label{eq:sinkhorn}
\begin{aligned}
&\bm{T}^{(k)}=\exp(\bm{a}^{(k)}\bm{1}_N^T + \bm{1}_D(\bm{b}^{(k)})^T +{\bm{X}}/{\alpha_0}),
\quad\bm{p}^{(k)} = \bm{T}^{(k)}\bm{1}_N,
\quad\bm{q}^{(k)} = (\bm{T}^{(k)})^T\bm{1}_D,\\
&\bm{a}^{(k+1)}=\frac{\alpha_1(\bm{a}^{(k)}+\alpha_0(\log\bm{p}_0-\log\bm{p}^{(k)}))}{\alpha_0(\alpha_0+\alpha_1)},
\quad\bm{b}^{(k+1)}=\frac{\alpha_2(\bm{b}^{(k)}+\alpha_0(\log\bm{q}_0-\log\bm{q}^{(k)}))}{\alpha_0(\alpha_0+\alpha_2)}.
\end{aligned}
\end{eqnarray}
$iii$) After $K$ steps, we obtain $\bm{P}_{\text{uot}}^*:=\bm{T}^{(K)}$. 
Applying the logarithmic stabilization strategy~\citep{chizat2018scaling,schmitzer2019stabilized}, we achieve the exponentiation and scaling in~(\ref{eq:sinkhorn}) by ``LogSumExp''.

The Sinkhorn-based UOTP layer unrolls the above iterative scheme by stacking $K$ Sinkhorn modules. 
Each module implements~(\ref{eq:sinkhorn}), which takes the dual variables as its input and updates them accordingly. 
The parameters include: $i$) \textbf{prior distributions $\{\bm{p}_0\in\Delta^{D-1}, \bm{q}_0\in\Delta^{N-1}\}$}, and $ii$) \textbf{module-specific weights $\{\bm{\alpha}_i=[\alpha_{i,k}]\in (0,\infty)^{K}\}_{i=0}^2$}, in which $\{\alpha_{i,k}\}_{i=0}^2$ are parameters of the $k$-th module. 
As shown in~\citep{sun2016deep,amos2017optnet}, introducing layer-specific parameters improves the model capacity. 
More details can be found at Appendix~\ref{app:sinkhorn}. 

\begin{algorithm}[t]
    \caption{UOTP$_{\text{BADMM}}(\bm{X};\{\bm{\alpha}_i\}_{i=0}^2,\bm{\rho},\bm{p}_0,\bm{q}_0)$}\label{alg:badmm_uotp}
	\begin{algorithmic}[1]
	    \STATE \textbf{Initialization:} Primal and auxiliary variable $\log\bm{P}^{(0)}=\log\bm{S}^{(0)}=\log(\bm{p}_0\bm{q}_0^T)$, $\log\bm{\mu}^{(0)}=\log\bm{p}_0$, $\log\bm{\eta}^{(0)}=\log\bm{q}_0$. Dual variables $\bm{Z}^{(0)}=\bm{0}_{D\times N}$, $\bm{z}_1^{(0)}=\bm{0}_D$, $\bm{z}_2^{(0)}=\bm{0}_N$.
	    \STATE{\textbf{For} $k=0,...,K-1$ ($K$ BADMM Modules)}
	        \STATE\quad\textbf{Update $\bm{P}$ by~(\ref{eq:T}) (Log-primal update):}\\
	        \quad\quad When applying the entropic regularizer, set $\bm{Y}=\log\bm{S}^{(k)} + {(\bm{X}-\bm{Z}^{(k)})}/{\rho_k}$, \\
            \quad\quad When applying the quadratic regularizer, set
            $\bm{Y}=\log\bm{S}^{(k)} + {(\bm{X}-\alpha_{0,k}\bm{S}^{(k)}-\bm{Z}^{(k)})}/{\rho_k}$,\\
            \quad\quad Update $\log\bm{P}^{(k+1)}=(\log\bm{\mu}^{(k)}-\text{LogSumExp}_{\text{col}}(\bm{Y}))\bm{1}_N^T + \bm{Y}$.
	        \STATE\quad\textbf{Update $\bm{S},\bm{\mu},\bm{\eta}$ by}~(\ref{eq:S}) \textbf{(Log-auxiliary update):}\\
	        \quad\quad When applying the entropic regularizer, set $\bm{Y}={(\bm{Z}^{(k)}+\rho_k\log\bm{P}^{(k+1)})}{(\alpha_{0,k}+\rho_k)}$, \\
            \quad\quad When applying the quadratic regularizer, set $\bm{Y}=\log\bm{P}^{(k+1)} + {(\bm{Z}^{(k)}-\alpha_{0,k}\bm{S}^{(k+1)})}{\rho_k}$, \\
            \quad\quad Update $\log\bm{S}^{(k+1)}=\bm{1}_D(\log\bm{\eta}^{(k)} - \text{LogSumExp}_{\text{row}}(\bm{Y}))^T + \bm{Y}$,\\
	        \quad\quad$\log\bm{\mu}^{(k+1)}=\frac{\rho_k \log\bm{\mu}^{(k)} + \alpha_{1,k}\log \bm{p}_0-\bm{z}_1^{(k)}}{\rho_k + \alpha_{1,k}}$, \quad$\log\bm{\eta}^{(k+1)}=\frac{\rho_k \log\bm{\eta}^{(k)}+\alpha_{2,k}\log\bm{q}_0-\bm{z}_2^{(k)}}{\rho_k + \alpha_{2,k}}$.
	        \STATE\quad\textbf{Update $\bm{Z},\bm{z}_1,\bm{z}_2$ (Dual update):} $\bm{Z}^{(k+1)}=\bm{Z}^{(k)}+\alpha_{0,k}(\bm{P}^{(k+1)}-\bm{S}^{(k+1)})$, \\
	        \quad\quad$\bm{z}_1^{(k+1)}=\bm{z}_1^{(k)}+\rho_k (\bm{\mu}^{(k+1)}-\bm{P}^{(k+1)}\bm{1}_N)$,\quad$\bm{z}_2^{(k+1)}=\bm{z}_2^{(k)}+\rho_k (\bm{\eta}^{(k+1)}-\bm{S}^{(k+1)T}\bm{1}_D)$.
	    \STATE\textbf{Output:} $\bm{P}^{*}:=\bm{P}^{(K)}$ and apply~(\ref{eq:uotp}) accordingly.
	\end{algorithmic}
\end{algorithm}

\subsection{Proposed Bregman ADMM-based UOTP layer}\label{ssec:badmm}
The Sinkhorn-based UOTP layer is restricted to solve the entropy-regularized UOT problem and may suffer from numerical instability issues, because the Sinkhorn scaling algorithm is designed for entropic optimal transport problems and is sensitive to the weight of the entropic regularizer~\citep{xie2020fast}. 
To extend the flexibility of model design and solve the numerical problem, we develop a new UOTP layer based on the Bregman ADMM algorithm~\citep{wang2014bregman,xu2020gromov}. 
Here, we rewrite~(\ref{eq:uot}) in an equivalent format by introducing three auxiliary variables $\bm{S}$, $\bm{\mu}$ and $\bm{\eta}$:
\begin{eqnarray}\label{eq:badmm1}
\begin{aligned}
    \sideset{}{_{\bm{P}=\bm{S},~\bm{P}\bm{1}_N=\bm{\mu},~\bm{S}^T\bm{1}_D=\bm{\eta}}}\min \langle -\bm{X},\bm{P}\rangle +\alpha_0\text{R}(\bm{P}, \bm{S}) + \alpha_1\text{KL}(\bm{\mu}|\bm{p}_0) + \alpha_2\text{KL}(\bm{\eta}|\bm{q}_0).
\end{aligned}
\end{eqnarray}
These three auxiliary variables correspond to the optimal transport $\bm{P}$ and its marginals. 
Here, the original smoothness regularizer $\text{R}(\bm{P})$ is rewritten based on the auxiliary variable $\bm{S}$.
When using the entropic regularizer, we can set $\text{R}(\bm{P}, \bm{S})=\langle \bm{S},\log\bm{S}-\bm{1}\rangle$.\footnote{Here, the regularizer's input is just $\bm{S}$, but we still denote it as $\text{R}(\bm{P}, \bm{S})$ for the consistency of notation.} 
When using the quadratic regularizer, we set $\text{R}(\bm{P},\bm{S})=\langle\bm{P},\bm{S}\rangle$. 
This problem can be further rewritten in a Bregman-augmented Lagrangian form by introducing three dual variables $\bm{Z}$, $\bm{z}_1$, $\bm{z}_2$ for the three constraints in~(\ref{eq:badmm1}), respectively. 
Accordingly, we solve the UOT problem by alternating optimization: At the $k$-th iteration, we rewrite~(\ref{eq:badmm1}) in the following the Bregman-augmented Lagrangian form for $\bm{P}$ and update $\bm{P}$ by
\begin{eqnarray}\label{eq:T}
\begin{aligned}
    \bm{P}^{(k+1)}=\arg\sideset{}{_{\bm{P}\in \Pi(\bm{\mu}^{(k)},\cdot)}}\min \langle -\bm{X},\bm{P}\rangle+\alpha_0\text{R}(\bm{P}, \bm{S}^{(k)}) +\langle \bm{Z}^{(k)},\bm{P}-\bm{S}^{(k)}\rangle + \rho\text{KL}(\bm{P}|\bm{S}^{(k)}).
\end{aligned}
\end{eqnarray}
Here, $\Pi(\bm{\mu}^{(k)},\cdot)=\{\bm{P}>\bm{0}|\bm{P}\bm{1}_N=\bm{\mu}^{(k)}\}$ is the one-side constraint, and $\sigma_{\text{row}}$ is a row-wise softmax operation. 
The KL-divergence term $\text{KL}(\bm{P}|\bm{S}^{(k)})$ is the Bregman divergence. 
Similarly, given $\bm{P}^{(k+1)}$, we update the auxiliary variables $\bm{S}$, $\bm{\mu}$ and $\bm{\eta}$ by
\begin{eqnarray}\label{eq:S}
\begin{aligned}
    \bm{S}^{(k+1)} &= \arg\sideset{}{_{\bm{S}\in\Pi(\cdot, \bm{\eta}^{(k)})}}\min \alpha_0\text{R}(\bm{P}^{(k+1)}, \bm{S}) + \langle \bm{Z}^{(k)}, \bm{P}^{(k+1)}-\bm{S}\rangle + \rho\text{KL}(\bm{S}|\bm{P}^{(k+1)}),\\
    \bm{\mu}^{(k+1)}&= \arg\sideset{}{_{\bm{\mu}}}\min \alpha_1\text{KL}(\bm{\mu}|\bm{p}_0) + \langle\bm{z}_1^{(k)}, \bm{\mu} -\bm{P}^{(k+1)}\bm{1}_{N}\rangle + \rho\text{KL}(\bm{\mu}| \bm{P}^{(k+1)}\bm{1}_{N}), \\
    \bm{\eta}^{(k+1)}&=\arg\sideset{}{_{\bm{\eta}}}\min
    \alpha_2\text{KL}(\bm{\eta}|\bm{q}_0) +  \langle\bm{z}_2^{(k)}, \bm{\eta} -(\bm{S}^{(k+1)})^T\bm{1}_{D}\rangle + \rho\text{KL}(\bm{\eta}| (\bm{S}^{(k+1)})^T\bm{1}_{D}),
\end{aligned}
\end{eqnarray}
All the optimization problems in~(\ref{eq:T}) and~(\ref{eq:S}) have closed-form solutions.
Finally, we update the dual variables as classic ADMM does.
More detailed derivation is given in Appendix~\ref{app:badmm}.

As shown in Figure~\ref{fig:badmm_uot} and Algorithm~\ref{alg:badmm_uotp}, our BADMM-based UOTP layer implements the above BADMM algorithm by stacking $K$ feed-forward computational modules. 
Each module updates the primal, auxiliary, and dual variables, in which the logarithmic stabilization strategy~\citep{chizat2018scaling,schmitzer2019stabilized} is applied.
Similar to the Sinkhorn-based UOTP layer, our BADMM-based UOTP layer also owns module-specific weights of regularizers and shared prior distributions. 
For the module-specific weights, besides the $\{\bm{\alpha}_i\}_{i=0}^2$, the BADMM-based UOTP layer contains one more vector $\bm{\rho}=[\rho_k]\in(0,\infty)^K$, $i.e.$, the weights of the Bregman divergence terms.

\subsection{Implementation details and comparisons}\label{ssec:implement}
\textbf{Reparametrization for unconstrained optimization.} 
The above UOTP layers have constrained parameters: $\{\bm{\alpha}_i\}_{i=0}^2$ and $\bm{\rho}$ are positive, $\bm{p}_0\in\Delta^{D-1}$, and $\bm{q}_0\in\Delta^{N-1}$.
We set $\{\bm{\alpha}_i=\text{softplus}(\bm{\beta}_i)\}_{i=0}^2$ and $\bm{\rho}=\text{softplus}(\bm{\tau})$, where $\{\bm{\beta}_i\}_{i=0}^{2}$ and $\bm{\tau}$ are unconstrained parameters. 
For the prior distributions, we can either fix them as uniform distributions, $i.e.$, $\bm{p}_0=\frac{1}{D}\bm{1}_D$ and $\bm{q}_0=\frac{1}{N}\bm{1}_N$, or implement them as learnable attention modules, $i.e.$, $\bm{p}_0=\text{softmax}(\bm{U}\bm{X}\bm{1}_N)$ and $\bm{q}_0=\text{softmax}(\bm{w}^T\text{tanh}(\bm{VX}))$~\citep{ilse2018attention}, where $\bm{U},\bm{V}\in\mathbb{R}^{D\times D}$ and $\bm{w}\in\mathbb{R}^D$ are unconstrained. 
As a result, our UOTP layers can be learned by stochastic gradient descent.

\begin{figure}[t]
\centering
    \subfigure[Approximations of various pooling layers]{
    	\includegraphics[height=3.5cm]{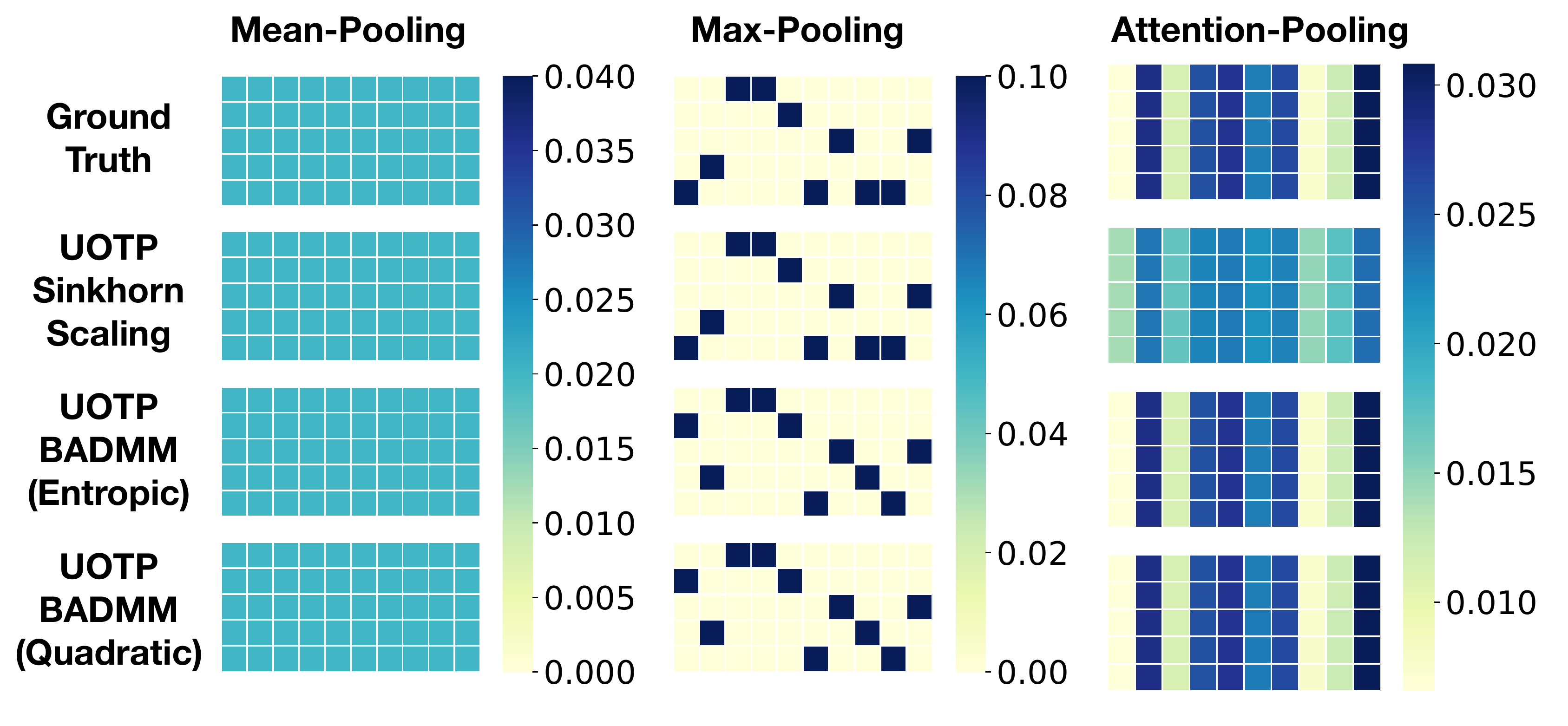}\label{fig:approx}
    }
    \subfigure[Comparisons on numerical stability]{
        \includegraphics[height=3.5cm]{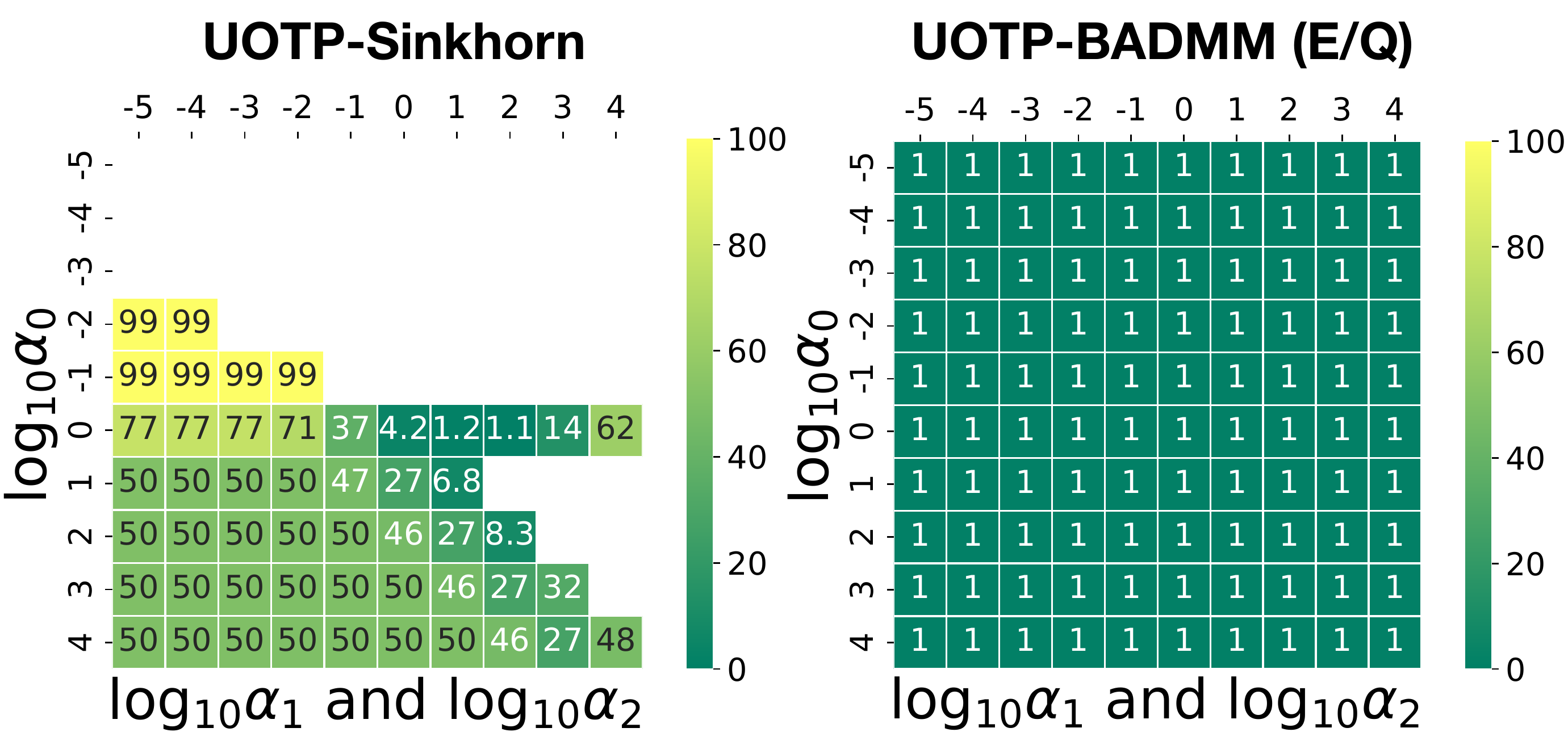}\label{fig:stability}
    }
    \vspace{-5pt}
    \caption{
        (a) Given an arbitrary $\bm{X}\in\mathbb{R}^{5\times 10}$, we approximate the $\bm{P}^*$'s corresponding to the mean-, max-, and attention-pooling operations. 
        In each subfigure, the matrices from top to bottom are the ground truth and the $\bm{P}^{*}$'s obtained by Sinkhorn-and BADMM-based UOTP layers, where $\alpha_0=\alpha_1=\alpha_2=10^4$ for mean-and attention-pooling, and $\alpha_0=\alpha_2=0.01$ and $\alpha_1=10^4$ for max-pooling.
        (b) Given $\bm{X}\in\mathbb{R}^{5\times 10}$, we learn $\bm{P}^*$'s under different configurations and calculate $\|\bm{P}^*\|_1$'s. 
        Each subfigure shows the $\|\bm{P}^*\|_1$'s, and the white regions correspond to \textsf{NaN}'s. 
        Our BADMM-based UOTP obtains the same numerical stability for both entropic and quadratic regularizers. 
    }
\end{figure}


\textbf{Precision of approximating conventional pooling methods.}
Proposition~\ref{prop:equiv} demonstrates that our UOTP layers can approximate, even be equivalent to, some existing pooling operations. 
We verify this proposition by the experimental results shown in Figure~\ref{fig:approx}. 
Under the configurations guided by Proposition~\ref{prop:equiv}, we use our UOTP layers to imitate mean-, max-, and attention-pooling operations. 
Both the Sinkhorn-based UOTP and the BADMM-based UOTP can reproduce the $\bm{P}^*$ of mean-pooling perfectly. 
The Sinkhorn-based UOTP achieves max-pooling with high precision, while the BADMM-based UOTP approximate max-pooling with some errors.
When approximating the attention-pooling, the BADMM-based UOTP works better than the Sinkhorn-based UOTP.

\textbf{Numerical stability.} 
We set $\alpha_1=\alpha_2$ and select $\alpha_0,\alpha_1,\alpha_2$ from $\{10^{-5},...,10^{4}\}$ for for each UOTP layer.
Accordingly, we derive 100 $\bm{P}^*$'s and check whether $\|\bm{P}^*\|_1=\sum_{d,n}|p_{dn}|\approx 1$ and whether $\bm{P}^*$ contains \textsf{NaN} elements. 
Figure~\ref{fig:stability} shows that the Sinkhorn-based UOTP merely works  under some configurations. 
Therefore, in the following experiments, we have to restrict the range of its parameters in some cases. 
Our BADMM-based UOTP owns better numerical stability, which avoids \textsf{NaN} elements and keeps $\|\bm{P}^*\|_1\approx 1$.

\begin{figure}[t]
    \centering
    \subfigure[Convergence comparison]{
    \includegraphics[height=5.5cm]{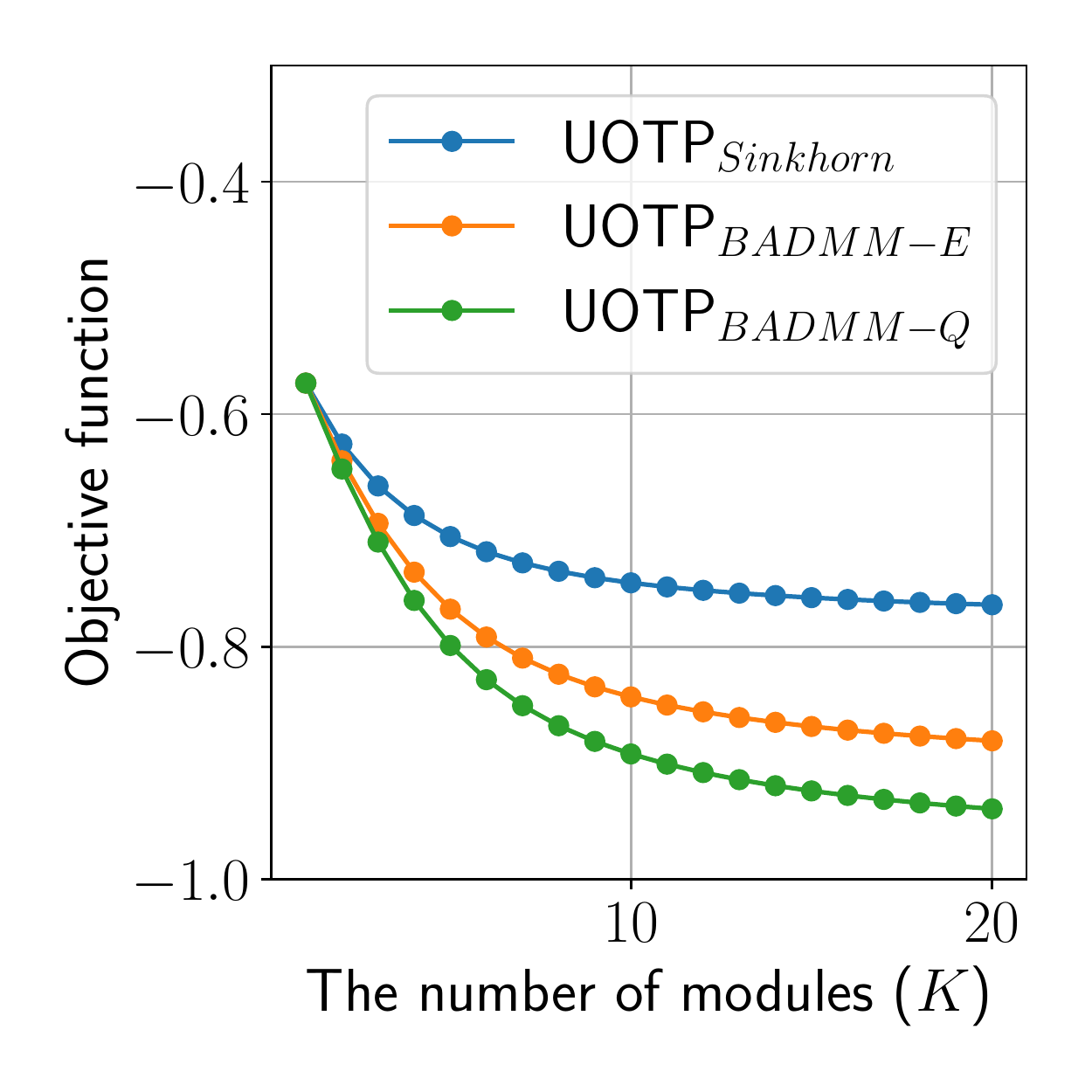}\label{fig:convergence}
    }
    \subfigure[Runtime comparison]{
    \includegraphics[height=5.5cm]{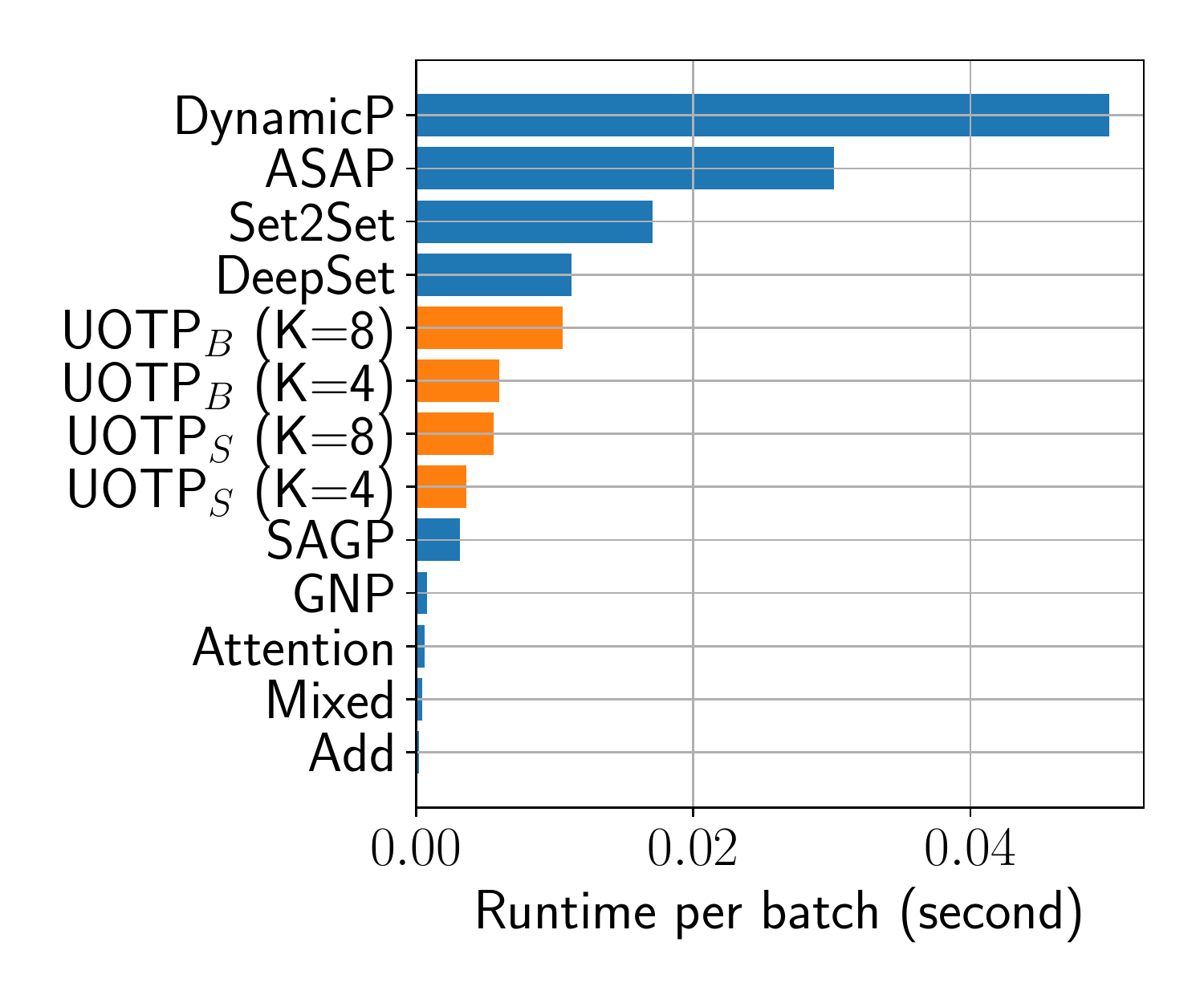}\label{fig:runtime}
    }
    \subfigure[Dynamics]{
    \includegraphics[height=5.5cm]{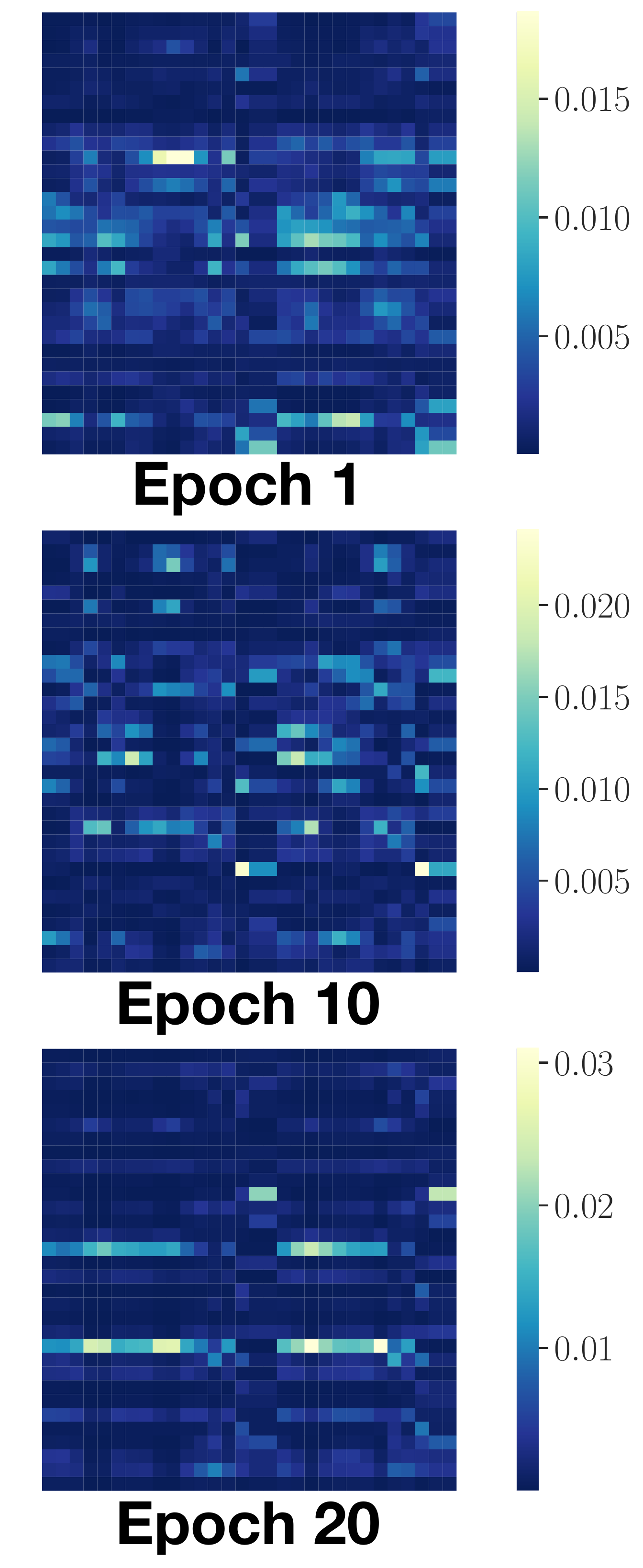}\label{fig:epoch}
    }
    \vspace{-5pt}
    \caption{
    Given a batch of $50$ sample sets, in which each sample set contains $500$, $100$-dimensional samples, we plot: 
    (a) The convergence of our UOTP layers with the increase of $K$;
    (b) the averaged feed-forward runtime of various pooling methods in $10$ trials on a single GPU (RTX 3090).
    (c) Given a batch of MUTAG graphs, we illustrate dynamics of the corresponding $\bm{P}^*$'s during training.}
\end{figure}

\textbf{Convergence and efficiency.} 
Given $N$ $D$-dimensional samples, the computational complexity of our UOTP layer is $\mathcal{O}(KND)$, where $K$ is the number of Sinkhorn/BADMM modules. 
As shown in Figure~\ref{fig:convergence}, with the increase of $K$, our UOTP layers reduce the objective of the UOT problem ($i.e.$, the expectation term $\langle -\bm{X},\bm{P}\rangle$ and its regularizers) consistently. 
When $K\geq 4$, the objective has been reduced significantly, and when $K\geq 8$, the objective has tended to convergent. 

Both the Sinkhorn-based and the BADMM-based UOTP layers involve two LogSumExp operations (the most time-consuming operations) per step. 
In practice, the BADMM-based UOTP may be slightly slower than the Sinkhorn-based UOTP in general --- it requires additional element-wise exponentiation to get $\bm{P},\bm{S},\bm{\mu},\bm{\eta}$ when updating dual variables (Line 5 of Algorithm~\ref{alg:badmm_uotp}). 
However, the runtime of our method is comparable to that of the learning-based pooling methods.
Figure~\ref{fig:runtime} shows the rank of various pooling methods on their runtime per batch. 
We can find that 
In particular, for the BADMM-based UOTP layer with $K=8$, its runtime is almost the same with that of DeepSet~\citep{zaheer2017deep}. 
For the Sinkhorn-based UOTP layer with $K=4$, its runtime is comparable to that of SAGP~\citep{lee2019self}. 
When setting $K\leq 8$, our UOTP layers are more efficient than the other pooling methods that stacks multiple computational modules ($e.g.$, Set2Set~\citep{vinyals2015order} and DynamicP~\citep{yan2018deep}).
According to the analysis above, in the following experiments, we set $K=4$ for our UOTP layers, which can achieve a trade-off between effectiveness and efficiency in most situations.

\section{Related Work}
\textbf{Pooling operations.}
Besides simple pooling operations, $e.g.$, mean/add-pooling, max-pooling, and their mixtures~\citep{lee2016generalizing}, learnable pooling layers, $e.g.$, Network-in-Network~\citep{lin2013network}, Set2Set~\citep{vinyals2015order}, DeepSet~\citep{zaheer2017deep}, and SetTransformer~\citep{lee2019set}, leverage multi-layer perceptrons, recurrent neural networks, and  transformers~\citep{vaswani2017attention} to achieve global pooling. 
The attention-pooling in~\citep{ilse2018attention} and the dynamic-pooling in~\citep{yan2018deep} merge multiple instances based on self-attentive mechanisms. 
Besides the above global pooling methods, some local pooling methods, $e.g.$, DiffPool~\citep{ying2018hierarchical}, SAGPooling~\citep{lee2019self}, and ASAPooling~\citep{ranjan2020asap}, are proposed for pooling graph-structured data. 
Recently, the OTK in~\citep{mialon2020trainable} and the WEGL in~\citep{kolouri2020wasserstein} consider the optimal transport between samples and achieve pooling operations for specific tasks. 
Different from above methods, our UOTP considers the optimal transport across sample index and feature dimension, which provides a new and generalized framework of global pooling.
Compared with the generalized norm-based pooling (GNP) in~\citep{ko2021learning}, our UOTP covers more pooling methods and can be interpreted well as an expectation-maximization strategy.

\textbf{Optimal transport-based machine learning.}
Optimal transport (OT) theory~\citep{villani2008optimal} has proven to be useful in machine learning tasks, $e.g.$, distribution matching~\citep{frogner2015learning,courty2016optimal}, data clustering~\citep{cuturi2014fast}, and generative modeling~\citep{arjovsky2017wasserstein,tolstikhin2018wasserstein}. 
The discrete OT problem is a linear programming problem~\citep{kusner2015word}. 
By adding an entropic regularizer~\citep{cuturi2013sinkhorn}, the problem becomes strictly convex and can be solved by the Sinkhorn scaling algorithm~\citep{sinkhorn1967concerning}. 
Along this direction, the stabilized Sinkhorn  algorithm~\citep{chizat2018scaling,schmitzer2019stabilized} and the proximal point method~\citep{xie2020fast} solve the entropic OT problem robustly. 
These algorithms can be extended to solve UOT problems~\citep{benamou2015iterative,pham2020unbalanced}.
Recently, some neural networks are designed to imitate the Sinkhorn-based algorithms, $e.g.$, the Gumbel-Sinkhorn network~\citep{mena2018learning}, the sparse Sinkhorn attention model~\citep{tay2020sparse}, the Sinkhorn autoencoder~\citep{patrini2020sinkhorn}, and the Sinkhorn-based transformer~\citep{sander2021sinkformers}. 
However, these models ignore the potentials of other algorithms, $e.g.$, the Bregman ADMM~\citep{wang2014bregman,xu2020gromov} and the smoothed semi-dual algorithm~\citep{blondel2018smooth}. 
None of them consider implementing global pooling layers as solving the UOT problem.

\begin{table*}[t]
\caption{Comparison on classification accuracy$\pm$Std. (\%) for different pooling layers.}\label{tab:mil+adgcl}
\resizebox{\textwidth}{!}{
\begin{threeparttable}
    \begin{tabular}{@{}c|c@{\hspace{4pt}}c@{\hspace{4pt}}c@{\hspace{4pt}}c|c@{\hspace{4pt}}c@{\hspace{4pt}}c@{\hspace{4pt}}c@{\hspace{4pt}}c@{\hspace{4pt}}c@{\hspace{4pt}}c@{\hspace{4pt}}c@{}}
    \hline\hline
    \multirow{2}{*}{Pooling}     &
    \multicolumn{4}{c|}{Multi-instance learning} &
    \multicolumn{8}{c}{Graph classification (ADGCL)} 
    \\
    \cline{2-13}
    &
    Messidor &
    Component       & 
    Function       & 
    Process     &
    NCII        & 
    PROTEINS     & 
    MUTAG       & 
    COLLAB       & 
    RDT-B       & 
    RDT-M5K        & 
    IMDB-B    & 
    IMDB-M   \\ 
    \hline
    Add  &  
    74.33$_{\pm\text{2.56}}$
    &
    \textbf{93.35$_{\pm\text{0.98}}$}
    &
    96.26$_{\pm\text{0.48}}$
    & 
    \textbf{97.41$_{\pm\text{0.21}}$}
    &
    67.96$_{\pm\text{0.43}}$
    & 
    72.97$_{\pm\text{0.54}}$
    & 
    \color{red}{\textbf{89.05$_{\pm\text{0.86}}$}}
    &  
    71.06$_{\pm\text{0.43}}$
    & 
    80.00$_{\pm\text{1.49}}$
    &  
    50.16$_{\pm\text{0.97}}$
    &
    70.18$_{\pm\text{0.87}}$  
    & 
    47.56$_{\pm\text{0.56}}$
    \\
    Mean & 
    74.42$_{\pm\text{2.47}}$
    &       
    93.32$_{\pm\text{0.99}}$
    &      
    96.28$_{\pm\text{0.66}}$
    &    
    97.20$_{\pm\text{0.14}}$
    &
    64.82$_{\pm\text{0.52}}$
    &
    66.09$_{\pm\text{0.64}}$
    & 
    86.53$_{\pm\text{1.62}}$
    &  
    72.35$_{\pm\text{0.44}}$
    & 
    83.62$_{\pm\text{1.18}}$
    &  
    \textbf{52.44$_{\pm\text{1.24}}$}
    &
    70.34$_{\pm\text{0.38}}$  
    & 
    48.65$_{\pm\text{0.91}}$ 
    \\
    Max  & 
    73.92$_{\pm\text{3.00}}$
    &   
    93.23$_{\pm\text{0.76}}$
    &       
    95.94$_{\pm\text{0.48}}$
    &      
    96.71$_{\pm\text{0.40}}$
    &
    65.95$_{\pm\text{0.76}}$
    &  
    72.27$_{\pm\text{0.33}}$
    & 
    85.90$_{\pm\text{1.68}}$
    &  
    73.07$_{\pm\text{0.57}}$
    & 
    82.62$_{\pm\text{1.25}}$
    &
    44.34$_{\pm\text{1.93}}$
    &
    70.24$_{\pm\text{0.54}}$
    & 
    47.80$_{\pm\text{0.54}}$  
    \\
    DeepSet  &  
    74.42$_{\pm\text{2.87}}$
    &     
    93.29$_{\pm\text{0.95}}$
    &  
    96.45$_{\pm\text{0.51}}$
    & 
    \color{red}{\textbf{97.64$_{\pm\text{0.18}}$}}
    &
    66.28$_{\pm\text{0.72}}$
    & 
    \color{red}{\textbf{73.76$_{\pm\text{0.47}}$}}
    & 
    87.84$_{\pm\text{0.71}}$
    &  
    69.74$_{\pm\text{0.66}}$
    & 
    82.91$_{\pm\text{1.37}}$
    &  
    47.45$_{\pm\text{0.54}}$
    &
    70.84$_{\pm\text{0.71}}$ 
    & 
    48.05$_{\pm\text{0.71}}$  
    \\
    Mixed  &  
    73.42$_{\pm\text{2.29}}$
    &
    \color{red}{\textbf{93.45$_{\pm\text{0.61}}$}}
    &   
    96.41$_{\pm\text{0.53}}$
    &      
    96.96$_{\pm\text{0.25}}$
    &
    66.46$_{\pm\text{0.74}}$
    &  
    72.25$_{\pm\text{0.45}}$
    & 
    87.30$_{\pm\text{0.87}}$
    & 
    \textbf{73.22$_{\pm\text{0.35}}$}
    & 
    \textbf{84.36$_{\pm\text{2.62}}$}
    & 
    46.67$_{\pm\text{1.63}}$
    &
    \textbf{71.28$_{\pm\text{0.26}}$} 
    & 
    48.07$_{\pm\text{0.25}}$  
    \\
    GatedMixed  &
    73.25$_{\pm\text{2.38}}$
    &    
    93.03$_{\pm\text{1.02}}$
    &    
    96.22$_{\pm\text{0.65}}$
    &      
    97.01$_{\pm\text{0.23}}$
    &
    63.86$_{\pm\text{0.76}}$
    &  
    69.40$_{\pm\text{1.93}}$
    & 
    87.94$_{\pm\text{1.28}}$
    &  
    71.94$_{\pm\text{0.40}}$
    & 
    80.60$_{\pm\text{3.89}}$
    &  
    44.78$_{\pm\text{4.53}}$  
    &
    70.96$_{\pm\text{0.60}}$
    & 
    48.09$_{\pm\text{0.44}}$
    \\
    Set2Set  & 
    73.58$_{\pm\text{3.74}}$
    &
    93.19$_{\pm\text{0.95}}$
    &   
    96.43$_{\pm\text{0.56}}$
    &    
    97.16$_{\pm\text{0.25}}$
    &
    65.10$_{\pm\text{1.12}}$
    &  
    68.61$_{\pm\text{1.44}}$
    & 
    87.77$_{\pm\text{0.86}}$
    &  
    72.31$_{\pm\text{0.73}}$
    & 
    80.08$_{\pm\text{5.72}}$
    &  
    49.85$_{\pm\text{2.77}}$
    &
    70.36$_{\pm\text{0.85}}$   
    & 
    48.30$_{\pm\text{0.54}}$
    \\
    Attention   & 
    74.25$_{\pm\text{3.67}}$
    &
    93.22$_{\pm\text{1.02}}$
    &   
    96.31$_{\pm\text{0.66}}$
    &    
    \textbf{97.24$_{\pm\text{0.16}}$}
    &
    64.35$_{\pm\text{0.61}}$
    & 
    67.70$_{\pm\text{0.95}}$
    & 
    88.08$_{\pm\text{1.22}}$
    &  
    72.57$_{\pm\text{0.41}}$
    & 
    81.55$_{\pm\text{4.39}}$
    &  
    51.85$_{\pm\text{0.66}}$
    &
    70.60$_{\pm\text{0.38}}$ 
    & 
    47.83$_{\pm\text{0.78}}$ 
    \\
    GatedAtt  &
    73.67$_{\pm\text{2.23}}$
    &
    \textbf{93.42$_{\pm\text{0.91}}$}
    &   
    \textbf{96.51$_{\pm\text{0.77}}$}
    &    
    97.18$_{\pm\text{0.14}}$
    &
    64.66$_{\pm\text{0.52}}$
    &  
    68.16$_{\pm\text{0.90}}$
    & 
    86.91$_{\pm\text{1.79}}$
    &  
    72.31$_{\pm\text{0.37}}$
    & 
    82.55$_{\pm\text{1.96}}$
    & 
    51.47$_{\pm\text{0.82}}$
    &
    70.52$_{\pm\text{0.31}}$  
    & 
    48.67$_{\pm\text{0.35}}$
    \\
    DynamicP      & 
    73.16$_{\pm\text{2.12}}$
    &
    93.26$_{\pm\text{1.30}}$
    &   
    \textbf{96.47$_{\pm\text{0.58}}$}
    &    
    97.03$_{\pm\text{0.14}}$
    &
    62.11$_{\pm\text{0.27}}$
    &  
    65.86$_{\pm\text{0.85}}$
    & 
    85.40$_{\pm\text{2.81}}$
    &  
    70.78$_{\pm\text{0.88}}$
    & 
    67.51$_{\pm\text{1.82}}$
    & 
    32.11$_{\pm\text{3.85}}$
    &
    69.84$_{\pm\text{0.73}}$ 
    &  
    47.59$_{\pm\text{0.48}}$
    \\
    GNP &
    73.54$_{\pm\text{3.68}}$
    &
    92.86$_{\pm\text{1.96}}$
    &   
    96.10$_{\pm\text{1.03}}$
    &    
    96.03$_{\pm\text{0.67}}$
    &
    \textbf{68.20$_{\pm\text{0.48}}$}
    &
    \textbf{73.44$_{\pm\text{0.61}}$}
    &
    \textbf{88.37$_{\pm\text{1.25}}$}
    &
    72.80$_{\pm\text{0.58}}$
    &
    81.93$_{\pm\text{2.23}}$
    &
    51.80$_{\pm\text{0.61}}$
    &
    70.34$_{\pm\text{0.83}}$
    &
    \textbf{48.85$_{\pm\text{0.81}}$}
    \\
    ASAP &
    ---
    &
    ---
    &
    ---
    &
    ---
    &
    \textbf{68.09$_{\pm\text{0.42}}$}
    &
    70.42$_{\pm\text{1.45}}$
    &
    87.68$_{\pm\text{1.42}}$
    &
    68.20$_{\pm\text{2.37}}$
    &
    73.91$_{\pm\text{1.50}}$
    &
    44.58$_{\pm\text{0.44}}$
    &
    68.33$_{\pm\text{2.50}}$
    &
    43.92$_{\pm\text{1.13}}$
    \\
    SAGP &
    ---
    &
    ---
    &
    ---
    &
    ---
    &
    67.48$_{\pm\text{0.65}}$
    &
    72.63$_{\pm\text{0.44}}$
    &
    87.88$_{\pm\text{2.22}}$
    &
    70.19$_{\pm\text{0.55}}$
    &
    74.12$_{\pm\text{2.86}}$
    &
    46.00$_{\pm\text{1.74}}$
    &
    70.34$_{\pm\text{0.74}}$
    &
    47.04$_{\pm\text{1.22}}$
    \\
    \hline
    UOTP$_{\text{Sinkhorn}}$  &  
    \color{red}{\textbf{75.42$_{\pm\text{2.96}}$}}
    &
    93.29$_{\pm\text{0.83}}$
    &   
    \color{red}{\textbf{96.62$_{\pm\text{0.48}}$}}
    &    
    97.08$_{\pm\text{0.11}}$
    &
    \color{red}{\textbf{68.27$_{\pm\text{1.06}}$}}
    &  
    \textbf{73.10$_{\pm\text{0.22}}$}
    & 
    \textbf{88.84$_{\pm\text{1.21}}$}
    &  
    71.20$_{\pm\text{0.55}}$
    
    & 
    81.54$_{\pm\text{1.38}}$
    &
    51.00$_{\pm\text{0.61}}$
    &
    70.74$_{\pm\text{0.80}}$
    &  
    47.87$_{\pm\text{0.43}}$
    \\
    UOTP$_{\text{BADMM-E}}$  &
    \textbf{74.83$_{\pm\text{2.07}}$}
    &
    93.16$_{\pm\text{1.02}}$
    &   
    96.17$_{\pm\text{0.43}}$
    &    
    97.15$_{\pm\text{0.16}}$
    &
    66.23$_{\pm\text{0.50}}$
    &  
    67.71$_{\pm\text{1.70}}$
    & 
    86.82$_{\pm\text{2.02}}$
    &
    \textbf{73.86$_{\pm\text{0.44}}$}
    & 
    \textbf{86.80$_{\pm\text{1.19}}$}
    &
    \textbf{52.25$_{\pm\text{0.75}}$}
    &
    \textbf{71.72$_{\pm\text{0.88}}$}
    &  
    \color{red}{\textbf{ 50.48$_{\pm\text{0.14}}$}}
    \\
    UOTP$_{\text{BADMM-Q}}$  &
    \textbf{75.08$_{\pm\text{2.06}}$}
    &
    93.13$_{\pm\text{0.94}}$
    &   
    96.09$_{\pm\text{0.46}}$
    &    
    97.08$_{\pm\text{0.17}}$
    &
    66.18$_{\pm\text{0.76}}$
    &  
    69.88$_{\pm\text{0.87}}$
    & 
    85.42$_{\pm\text{1.10}}$
    &  
    \color{red}{\textbf{74.14$_{\pm\text{0.24}}$}}  
    & 
    \color{red}{\textbf{87.72$_{\pm\text{1.03}}$}}
    &
    \color{red}{\textbf{52.79$_{\pm\text{0.60}}$}}
    &
    \color{red}{\textbf{72.34$_{\pm\text{0.50}}$}}
    &  
    \textbf{49.36$_{\pm\text{0.52}}$}
    \\ 
    \hline\hline
    \end{tabular}
    \begin{tablenotes}
    \item[*] The top-3 results of each data are bolded and the best result is in red.
    \end{tablenotes}
\end{threeparttable}
}
\end{table*}

\section{Experiments}\label{sec:exp}
In principle, applying our UOTP layers can reduce the difficulty of the design and selection of global pooling --- after learning based on observed data, our UOTP layers may either imitate some existing global pooling methods or lead to some new pooling layers fitting the data better.
To verify this claim, we test our UOTP layers (\textbf{UOTP$_{\text{Sinkhorn}}$} and \textbf{UOTP$_{\text{BADMM-E}}$} with the entropic regularizer, and \textbf{UOTP$_{\text{BADMM-Q}}$} with the quadratic regularizer) in three tasks, $i.e.$, multi-instance learning, graph classification, and image classification. 
The baselines include $i)$ classic \textbf{Add}-Pooling, \textbf{Mean}-Pooling, and \textbf{Max}-Pooling; $ii)$ the \textbf{Mixed}-Pooling and the \textbf{GatedMixed}-Pooling in~\citep{lee2016generalizing}; $iii)$ the learnable pooling layers like \textbf{DeepSet}~\citep{zaheer2017deep}, \textbf{Set2Set}~\citep{vinyals2015order}, \textbf{DynamicP}~\citep{yan2018deep}, \textbf{GNP}~\citep{ko2021learning}, and the \textbf{Attention}-Pooling and \textbf{GatedAtt}ention-Pooling in~\citep{ilse2018attention};
and $iv)$ \textbf{SAGP}~\citep{lee2019self} and \textbf{ASAP}~\citep{ranjan2020asap} for graph pooling.
We ran our experiments on a server with two  RTX3090 GPUs. 
\textbf{Experimental results and  implementation details are shown below and in Appendix~\ref{app:exp}.}

\textbf{Multi-instance learning.}
We consider four MIL tasks, which correspond to a disease diagnose dataset (Messidor~\citep{decenciere2014feedback}) and three gene ontology categorization datasets (Component, Function, and Process~\citep{blaschke2005evaluation}).
For each dataset, we learn a bag-level classifier, which embeds a bag of instances as input, merges the instances' embeddings via pooling, and finally, predicts the bag's label by a classifier. 
We use the AttentionDeepMIL in~\citep{ilse2018attention}, a representative bag-level classifier, as the backbone model and plug different pooling layers into it. 

\textbf{Graph classification.}
We consider eight representative graph classification datasets in the TUDataset~\citep{Morris+2020}, including three biochemical molecule datasets (NCII, MUTAG, and PROTEINS) and five social network datasets (COLLAB, RDT-B, RDT-M5K, IMDB-B, and IMDB-M). 
For each dataset, we implement the adversarial graph contrastive learning method (ADGCL)~\citep{suresh2021adversarial}, learning a graph isomorphism network (GIN)~\citep{xu2018powerful} to obtain graph embeddings. 
We apply different pooling operations to the GIN and use the learned graph embeddings to train an SVM classifier. 

Table~\ref{tab:mil+adgcl} presents the averaged classification accuracy and the standard deviation achieved by different methods under 5-fold cross-validation. 
For the multi-instance learning tasks, the performance of the UOTP layers is at least comparable to that of the baselines.
For the graph classification tasks, our BADMM-based UOTP layers even achieve the best performance on five social network datasets. 
These results indicate that our work simplifies the design and selection of global pooling to some degree.
In particular, none of the baselines perform consistently well across all the datasets, while our UOTP layers are comparable to the best baselines in most situations, whose performance is more stable and consistent. 
Therefore, in many learning tasks, instead of testing various global pooling methods empirically, we just need to select an algorithm ($i.e.$, Sinkhorn-scaling or Bregman ADMM) to implement the UOTP layer, which can achieve encouraging performance.

\begin{figure}[t]
    \centering
    \subfigure[MUTAG]{
    \includegraphics[height=5.3cm]{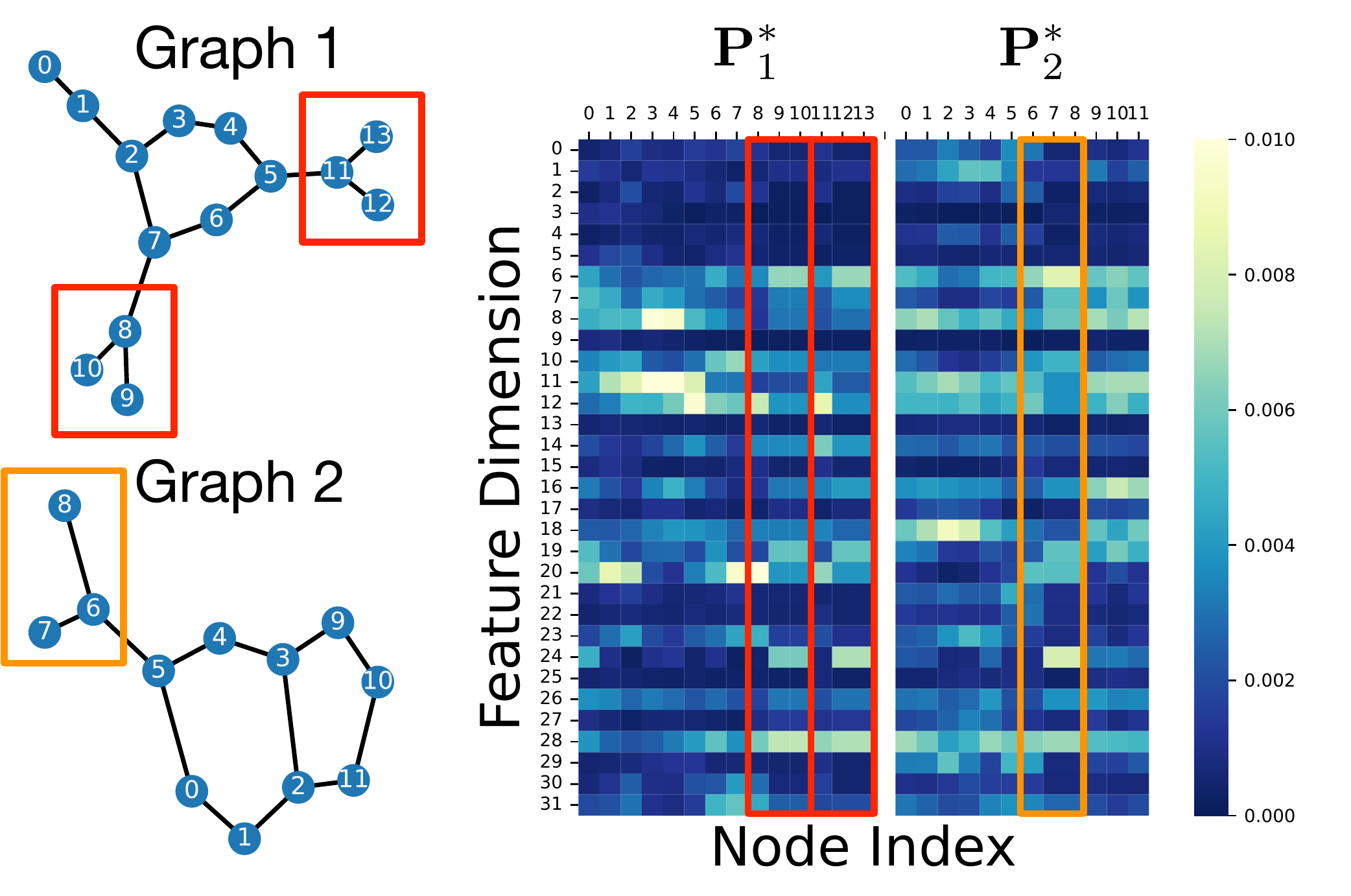}\label{fig:ration1}
    }
    \subfigure[IMDB-B]{
    \includegraphics[height=5.3cm]{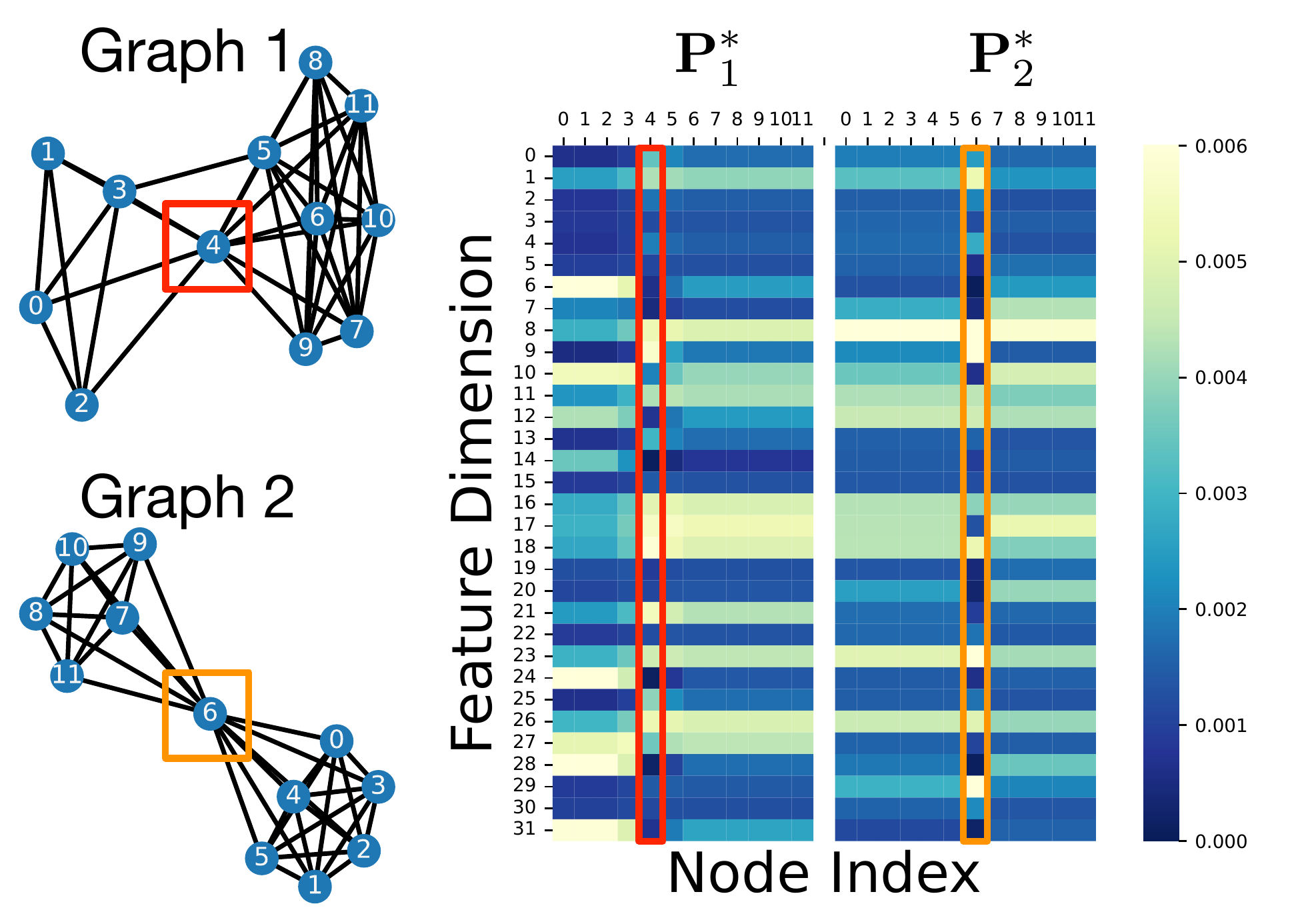}\label{fig:ration2}
    }
    \vspace{-5pt}
    \caption{
    (a) The visualizations of two MUTAG graphs and their $\bm{P}^*$'s. 
    For the ``V-shape'' subgraphs, their submatrices in the $\bm{P}^*$'s are marked by color frames.
    (b) The visualizations of two IMDB-B graphs and their $\bm{P}^*$'s. 
    For each graph, its key node connecting two communities and the corresponding column in the $\bm{P}^*$'s are marked by color frames.}\label{fig:ration}
\end{figure}

\textbf{Dynamics and rationality.}
Take the UOTP$_{\text{BADMM-E}}$ layer used for the MUTAG dataset as an example. 
For a validation batch, we visualize the dynamics of the corresponding $\bm{P}^*$'s in different epochs in Figure~\ref{fig:epoch}. 
In the beginning, the $\bm{P}^*$ is relatively dense because the node embeddings are not fully trained and may not be distinguishable.
With the increase of epochs, the $\bm{P}^*$ becomes sparse and focuses more on significant sample-feature pairs. 
Additionally, to verify the rationality of the learned $\bm{P}^*$, we visualize some graphs and their $\bm{P}^*$'s in Figure~\ref{fig:ration}.
For the ``V-shape'' subgraphs in the two MUTAG graphs, we compare the corresponding submatrices shown in their $\bm{P}^*$'s. 
These submatrices obey the same pattern, which means that for the subgraphs shared by different samples, the weights of their node embeddings will be similar. 
For the key nodes in the two IMDB-B graphs, their corresponding columns in the $\bm{P}^*$'s are distinguished from other columns. 
For the nodes belonging to different communities, their columns in the $\bm{P}^*$'s own significant clustering structures.

\begin{table}[t]
    \caption{Comparisons for ResNets and our ResNets + UOTP on validation accuracy (\%)}
    \label{tab:resnet}
    \centering
    \resizebox{\textwidth}{!}{
    \begin{threeparttable}   
    \begin{small}
    \begin{tabular}{c|c|ccccc}
    \hline\hline
                   &Learning Strategy
                   &ResNet18
                   &ResNet34
                   &ResNet50
                   &ResNet101
                   &ResNet152\\
    \hline
    \multirow{2}{*}{Top-5}
    &100 Epochs (A2DP)
    &89.084
    &91.433
    &92.880
    &93.552 
    &94.048 \\
    &90 Epochs (A2DP) + 10 Epochs (UOTP)
    &\textbf{89.174}
    &\textbf{91.458}
    &\textbf{93.006}
    &\textbf{93.622}
    &\textbf{94.060}\\
    \hline
    \multirow{2}{*}{Top-1}
    &100 Epochs (A2DP)
    &69.762
    &73.320
    &76.142
    &77.386
    &78.324\\
    &90 Epochs (A2DP) + 10 Epochs (UOTP)
    &\textbf{69.906}
    &\textbf{73.426}
    &\textbf{76.446}
    &\textbf{77.522}
    &\textbf{78.446}\\
    \hline\hline
    \end{tabular}
    \end{small}
    \end{threeparttable}  
    }
\end{table}

\textbf{Image classification.}
Given a ResNet~\citep{he2016deep}, we replace its ``adaptive 2D mean-pooling layer (A2DP)'' with our UOTP$_{\text{BADMM-E}}$ layer and finetune the modified model on ImageNet~\citep{deng2009imagenet}. 
In particular, given the output of the last convolution layer of the ResNet, $i.e.$, $\bm{X}_{\text{in}}\in \mathbb{R}^{B\times C\times H\times W}$, our UOTP layer fuses the data and outputs $\bm{X}_{\text{out}}\in \mathbb{R}^{B\times C\times 1\times 1}$. 
In this experiments, we apply a two-stage learning strategy: we first train a ResNet in 90 epochs; and then we replace its A2DP layer with our UOTP layer; finally, we fix other layers and train our UOTP layer in 10 epochs.
The learning rate is 0.001, and the batch size is 256. 
Table~\ref{tab:resnet} shows that using our UOTP layer helps to improve the classification accuracy and the improvement is consistent for different ResNets. 

\textbf{Limitations and future work.} 
The improvements in~Table~\ref{tab:resnet} are incremental because we just replace a single global pooling layer with our UOTP layer. 
When training the ResNets with UOTP layers from scratch, the improvements are not so  significant, either --- after training ResNet18+UOTP with 100 epochs, the top-1 accuracy is 69.920\% and the top-5 accuracy is 89.198\%. 
In principle, replacing more local pooling layers with our UOTP layers may bring better performance. 
However, given a tensor $\bm{X}_{\text{in}}\in \mathbb{R}^{B\times C\times H\times W}$, a local pooling merges each patch with size $(B\times C\times 2\times 2)$ into $B$ $C$-dimensional vectors and outputs $\bm{X}_{\text{out}}\in \mathbb{R}^{B\times C\times \frac{H}{2}\times \frac{W}{2}}$, which involves $\frac{BHW}{4}$ pooling operations.  
Such a local pooling requires an efficient CUDA implementation of the UOTP layers, which will be our future work. 

\section{Conclusion}
In this work, we studied global pooling through the lens of optimal transport and demonstrated that many existing global pooling operations correspond to solving a UOT problem with different configurations. 
We implemented the UOTP layer based on different algorithms and analyzed their stability and complexity in details.
Experiments verify their feasibility in various learning tasks. 

\clearpage
\newpage
\bibliography{refs}
\bibliographystyle{iclr2023_conference}

\clearpage
\newpage
\appendix
\section{Delayed Proofs}\label{app:proof}
\subsection{Proof of Theorem~\ref{thm:pi} and Corollary~\ref{coro:pi}}\label{app:thm}
\begin{proof}
Suppose that $\bm{q}_{0}$ be a permutation-equivariant function of $\bm{X}$, $i.e.$, $\bm{q}_{0}=g(\bm{X})$, where $g:\mathcal{X}_D\mapsto \Delta^{N-1}$ and $\bm{q}_{0,\pi}=g_{\pi}(\bm{X})=g(\bm{X}_{\pi})$ for an arbitrary permutation $\pi: \{1,...,N\}\mapsto\{1,....,N\}$. 
In such a situation, if $\bm{P}^*$ is the optimal solution of~(\ref{eq:uot}) given $\bm{X}$, then $\bm{P}^*_{\pi}$ must be the optimal solution of~(\ref{eq:uot}) given $\bm{X}_{\pi}$ because for each term in~(\ref{eq:uot}), we have
\begin{eqnarray}
\begin{aligned}
   &\langle-\bm{X},\bm{P}\rangle= \langle-\bm{X}_{\pi},\bm{P}_{\pi}\rangle,\\
   &\text{R}(\bm{P})=\text{R}(\bm{P}_{\pi})~\text{for both entropic and quadratic cases},\\
   &\text{KL}(\bm{P1}_N|\bm{p}_0)=\text{KL}(\bm{P}_{\pi}\bm{1}_N|\bm{p}_0),~\text{and}\\
   &\text{KL}(\bm{P}^T\bm{1}_D|\bm{q}_0)=\text{KL}(\bm{P}_{\pi}^T\bm{1}_D|\bm{q}_{0,\pi})=\text{KL}(\bm{P}_{\pi}^T\bm{1}_D|g(\bm{X}_{\pi})).
\end{aligned}
\end{eqnarray}
As a result, $\bm{P}^*$ is also a permutation-equivariant function of $\bm{X}$, $i.e.$, $\bm{P}^*_{\pi}(\bm{X})=\bm{P}^*(\bm{X}_{\pi})$, and accordingly, we have
\begin{eqnarray}
\begin{aligned}
f_{\text{uot}}(\bm{X}_{\pi})
=&(\bm{X}_{\pi}\odot (\text{diag}^{-1}( \bm{P}^*(\bm{X}_{\pi})\bm{1}_N)\bm{P}^*(\bm{X}_{\pi})))\bm{1}_N\\
=&(\bm{X}_{\pi}\odot (\text{diag}^{-1}( \bm{P}_{\pi}^*(\bm{X})\bm{1}_N)\bm{P}_{\pi}^*(\bm{X})))\bm{1}_N\\
=&(\bm{X}_{\pi}\odot (\text{diag}^{-1}( \bm{P}^*(\bm{X})\bm{1}_N)\bm{P}_{\pi}^*(\bm{X})))\bm{1}_N\\
=&(\bm{X}\odot (\text{diag}^{-1}( \bm{P}^*(\bm{X})\bm{1}_N)\bm{P}^*(\bm{X})))\bm{1}_N\\
=&f_{\text{uot}}(\bm{X}),
\end{aligned}
\end{eqnarray}
which completes the proof.

\textbf{Proof of Corollary~\ref{coro:pi}:} 
When $\bm{q}_0$ is uniform, we have $\text{KL}(\bm{P}^T\bm{1}_D|\bm{q}_0)=\text{KL}(\bm{P}_{\pi}^T\bm{1}_D|\bm{q}_{0,\pi})$, which provides a special case satisfying the condition shown in Theorem~\ref{thm:pi}. 
Accordingly, this setting also makes the optimal solution $\bm{P}^*$ permutation-equivariant to $\bm{X}$ and leads to the derivation in Theorem~\ref{thm:pi}.
\end{proof}

\subsection{Proof of Propositions~\ref{prop:equiv} and~\ref{prop:mix}}\label{app:prop}
\begin{proof}
\textbf{Equivalence to mean-pooling:}
For~(\ref{eq:uot}), when $\alpha_1,\alpha_2\rightarrow\infty$, $\bm{p}_0=\frac{1}{D}\bm{1}_D$ and $\bm{q}_0=\frac{1}{N}\bm{1}_N$, we require the marginals of $\bm{P}^*$ to match with $\bm{p}_0$ and $\bm{q}_0$ strictly. 
Additionally, $\alpha_0\rightarrow\infty$ means that the first term $\langle -\bm{X},\bm{P}\rangle$ becomes ignorable compared to the second term $\alpha_0\text{R}(\bm{P})$. 
Therefore, the unbalanced optimization problem in~(\ref{eq:uot}) degrades to the following minimization problem:
\begin{eqnarray}
\bm{P}^*=\arg\sideset{}{_{\bm{P}\in \Pi(\frac{1}{D}\bm{1}_D, \frac{1}{N}\bm{1}_N)}}\min \text{R}(\bm{P}).
\end{eqnarray}
When $\text{R}(\bm{P})$ is the entropic or the quadratic regularizer, the objective function is strictly-convex, and the optimal solution is $\bm{P}^*=[\frac{1}{DN}]$. 
Therefore, the corresponding $f_{\text{uot}}$ becomes the mean-pooling operation.

\textbf{Equivalence to max-pooling:}
For~(\ref{eq:uot}), when $\alpha_0=\alpha_2\rightarrow 0$, both the entropic term and the KL-based regularizer on $\bm{P}^T\bm{1}_D$ are ignorable. 
Additionally, $\alpha_1\rightarrow\infty$ and $\bm{p}_0=\frac{1}{D}\bm{1}_D$ mean that $\bm{P1}_N=\frac{1}{D}\bm{1}_D$ strictly.
The problem in~(\ref{eq:uot}) becomes
\begin{eqnarray}
\bm{P}^*=\arg\sideset{}{_{\bm{P}\in \Pi(\frac{1}{D}\bm{1}_D, \cdot)}}\max \langle\bm{X},\bm{P}\rangle,
\end{eqnarray}
whose optimal solution obviously corresponds to setting $p^*_{dn}=\frac{1}{D}$ if and only if $n=\arg\max_{m}\{x_{dm}\}_{m=1}^M$.
Therefore, the corresponding $f_{\text{uot}}$ becomes the max-pooling operation.

\textbf{Equivalence to attention-pooling:}
Similar to the case of mean-pooling, under such a configuration, the problem in~(\ref{eq:uot}) becomes the following minimization problem:
\begin{eqnarray}
\bm{P}^*=\arg\sideset{}{_{\bm{P}\in \Pi(\frac{1}{D}\bm{1}_D, \bm{a}_{\bm{X}})}}\max \text{R}(\bm{P}).
\end{eqnarray}
Similar to the case of mean-pooling, when $\text{R}(\bm{P})$ is the entropic or the quadratic regularizer, the objective function is strictly-convex, and the optimal solution is $\bm{P}^*=\frac{1}{D}\bm{1}_D\bm{a}_{\bm{X}}^T$. 
Accordingly, the corresponding $f_{\text{uot}}$ becomes the self-attentive pooling operation.

\textbf{Equivalence to mixed mean-max pooling:} For the mixed mean-max pooling, we have
\begin{eqnarray}\label{eq:hot}
\begin{aligned}
f_{\text{mix}}(\bm{X})&=\omega\text{MeanPool}(\bm{X})+(1-\omega)\text{MaxPool}(\bm{X})\\
&=\omega f_{\text{uot}}(\bm{X};\bm{\theta}_1) + (1-\omega)f_{\text{uot}}(\bm{X};\bm{\theta}_2)
=\underbrace{[f_{\text{uot}}(\bm{X};\bm{\theta}_1),f_{\text{uot}}(\bm{X};\bm{\theta}_2)]}_{\bm{Y}\in\mathbb{R}^{D\times 2}}[\omega, 1-\omega]^T\\
&=\Bigl(\bm{Y}\odot \text{diag}^{-1}\bigl( \underbrace{\overbrace{\tfrac{1}{D}\bm{1}_D}^{\bm{p}_0}\overbrace{[\omega,1-\omega]}^{\bm{q}_0^T}}_{\bm{P}^*}\bm{1}_2\bigr)\bigl(\tfrac{1}{D}\bm{1}_D[\omega,1-\omega]\bigr)\Bigr)\bm{1}_2
=f_{\text{uot}}(\bm{Y};\bm{\theta}_3).
\end{aligned}
\end{eqnarray}
Here, the first equation is based on Proposition~\ref{prop:equiv} --- we can replace $\text{MeanPool}(\bm{X})$ and $\text{MaxPool}(\bm{X})$ with $f_{\text{uot}}(\bm{X};\bm{\theta}_1)$ and $f_{\text{uot}}(\bm{X};\bm{\theta}_2)$, respectively, where $\bm{\theta}_1=\{\infty,\infty,\infty,\frac{1}{D}\bm{1}_D,\frac{1}{N}\bm{1}_N\}$ and $\bm{\theta}_2=\{0, \infty,0,\frac{1}{D}\bm{1}_D,-\}$. 
The concatenation of $f_{\text{uot}}(\bm{X};\bm{\theta}_1)$ and $f_{\text{uot}}(\bm{X};\bm{\theta}_2)$ is a matrix with size $D\times 2$, denoted as $\bm{Y}$. 
As shown in the third equation of~(\ref{eq:hot}), the $f_{\text{mix}}(\bm{X})$ in~(\ref{eq:mmp}) can be rewritten based on $\bm{p}_0=\frac{1}{D}\bm{1}_D$, $\bm{q}_0=[\omega,1-\omega]^T$, and the rank-1 matrix $\bm{P}^*=\bm{p}_0\bm{q}_0^T$.
The formulation corresponds to passing $\bm{Y}$ through the third ROTP operation, $i.e.$, $f_{\text{uot}}(\bm{Y};\bm{\theta}_3)$, where $\bm{\theta}_3=\{\infty,\infty,\infty,\frac{1}{D}\bm{1}_D,[\omega,1-\omega]^T\}$. 
\end{proof}

\section{The Details of Sinkhorn Scaling for UOT Problem}\label{app:sinkhorn}
\subsection{The dual form of UOT problem}
In the case of using the entropic regularizer, given the prime form of the UOT problem in~(\ref{eq:uot}), we can formulate its dual form as
\begin{eqnarray}\label{eq:dual1}
\begin{aligned}
    \sideset{}{_{\bm{a}\in\mathbb{R}^D,\bm{b}\in\mathbb{R}^N}}\min \alpha_0\sideset{}{_{d,n=1}^{D,N}}\sum \exp\bigl( \frac{a_d+b_n + x_{dn}}{\alpha_1} \bigr)+F^*(-\bm{a}) + G^*(-\bm{b}),
\end{aligned}
\end{eqnarray}
where
\begin{eqnarray}\label{eq:dual2}
\begin{aligned}
    F^*(\bm{a})&=\sideset{}{_{\bm{z}\in\mathbb{R}^D}}\max\bm{z}^T\bm{a}-\alpha_1 \text{KL}(\bm{z} | \bm{p}_0)
    =\alpha_1\langle\exp\bigl(\frac{1}{\alpha_1}\bm{a}\bigr) - \bm{1}_D, \bm{p}_0\rangle.\\
    G^*(\bm{b})&=\sideset{}{_{\bm{z}\in\mathbb{R}^N}}\max\bm{z}^T\bm{b}-\alpha_2 \text{KL}(\bm{z} | \bm{q}_0)
    =\alpha_2\langle\exp\bigl(\frac{1}{\alpha_2}\bm{b}\bigr) - \bm{1}_N, \bm{q}_0\rangle.
\end{aligned}
\end{eqnarray}
Plugging~(\ref{eq:dual2}) into~(\ref{eq:dual1}) leads to the dual form in~(\ref{eq:dual3}):
\begin{eqnarray}\label{eq:dual3}
\begin{aligned}
    \sideset{}{_{\bm{a}\in\mathbb{R}^D,\bm{b}\in\mathbb{R}^N}}\min &\alpha_0\sideset{}{_{d,n=1}^{D,N}}\sum \exp\left( \frac{a_d+b_n + x_{dn}}{\alpha_0} \right)+\\
    &\alpha_1\langle \exp(-\frac{1}{\alpha_1}\bm{a}), \bm{p}_0\rangle +
    \alpha_2\langle \exp(-\frac{1}{\alpha_2}\bm{b}), \bm{q}_0\rangle.
\end{aligned}
\end{eqnarray}
This problem can be solved by the iterative steps shown in~(\ref{eq:sinkhorn}).

\begin{algorithm}[t]
    \caption{UOTP$_{\text{Sinkhorn}}(\bm{X};\{\bm{\alpha}_i\}_{i=0}^2,\bm{p}_0,\bm{q}_0)$}\label{alg:sinkhorn_uotp}
    \label{alg:sinkhorn}
	\begin{algorithmic}[1]
	    \STATE Initialize $\bm{a}^{(0)}=\bm{0}_D$ and $\bm{b}^{(0)}=\bm{0}_N$, $\bm{Y}^{(0)}=\frac{1}{\alpha_{0,0}}\bm{X}$.
	    \STATE {\textbf{For} $k=0,...,K-1$ ($K$ Sinkhorn Modules)}
	        \STATE\quad $\log\bm{p} =\text{LogSumExp}_{\text{col}}(\bm{Y}^{(k)})$,\quad $\log\bm{q} =\text{LogSumExp}_{\text{row}}(\bm{Y}^{(k)})$.
	        \STATE\quad $\bm{a}^{(k+1)}=\frac{\alpha_{1,k}(\bm{a}^{(k)}+\alpha_{0,k}(\log\bm{p}_0-\log\bm{p}))}{\alpha_{0,k}(\alpha_{0,k}+\alpha_{1,k})}$,\quad  $\bm{b}^{(k+1)}=\frac{\alpha_{2,k}(\bm{b}^{(k)}+\alpha_{0,k}(\log\bm{q}_0-\log\bm{q}))}{\alpha_{0,k}(\alpha_{0,k}+\alpha_{2,k})}$.
	        \STATE\quad Logarithmic Scaling:~$\bm{Y}^{(k+1)}=\frac{1}{\alpha_{0,k}}\bm{X} + \bm{a}^{(k+1)}\bm{1}_N^T + \bm{1}_D(\bm{b}^{(k+1)})^T$.
	    \STATE\textbf{Output:} $\bm{P}^*:=\exp(\bm{Y}^{(K)})$ and apply~(\ref{eq:uotp}) accordingly.
	\end{algorithmic}
\end{algorithm}

\subsection{The implementation of the Sinkhorn-based UOTP layer}
As shown in Algorithm~\ref{alg:sinkhorn_uotp}, the Sinkhorn-based UOTP layer unrolls the iterative Sinkhorn scaling by stacking $K$ modules. 
The Sinkhorn-based UOTP layer unrolls the above iterative scheme by stacking $K$ modules. 
Each module implements~(\ref{eq:sinkhorn}), which takes the dual variables as its input and updates them accordingly.

\section{The Details of Bregman ADMM for UOT Problem}\label{app:badmm}
For the UOT problem with auxiliary variables ($i.e.$,~(\ref{eq:badmm1})), we can write its Bregman augmented Lagrangian form as
\begin{align}\label{eq:bregmanAL}
&\sideset{}{_{\bm{P},\bm{S},\bm{\mu},\bm{\eta},\bm{Z},\bm{z}_1,\bm{z}_2}}\min\underbrace{\langle-\bm{X},\bm{P}\rangle}_{\text{OT problem}} +\underbrace{\alpha_0\text{R}(\bm{P},\bm{S})}_{\text{Regularizer 1}} + \underbrace{\alpha_1\text{KL}(\bm{\mu}|\bm{p}_0)}_{\text{Regularizer 2}} +\underbrace{ \alpha_2\text{KL}(\bm{\eta}|\bm{q}_0)}_{\text{Regularizer 3}}+\\
&\underbrace{\overbrace{\langle\bm{Z},\bm{P}-\bm{S}\rangle + \rho\text{Div}(\bm{P},\bm{S})}^{\text{Constraint 1, for $\bm{T}$ and $\bm{S}$}} + \overbrace{\langle\bm{z}_1,\bm{\mu}-\bm{P}\bm{1}_N\rangle +
\rho\text{Div}(\bm{\mu},\bm{P}\bm{1}_N)}^{\text{Constraint 2, for $\bm{\mu}$ and $\bm{T}$}} +
\overbrace{\langle\bm{z}_2,\bm{\eta}-\bm{S}^T\bm{1}_D\rangle +
\rho\text{Div}(\bm{\eta},\bm{S}^T\bm{1}_N)}^{\text{Constraint 3, for $\bm{\eta}$ and $\bm{S}$}}}_{\text{Bregman augmented Lagrangian terms}}.\nonumber
\end{align}
Here, $\text{Div}(\cdot,\cdot)$ represents the Bregman divergence term, which is implemented as the KL-divergence in this work.
The second line of~(\ref{eq:bregmanAL}) contains the Bregman augmented Lagrangian terms, which correspond to the three constraints in~(\ref{eq:badmm1}). 

At the $k$-th iteration, given current variables $\{\bm{P}^{(k)},\bm{S}^{(k)},\bm{\mu}^{(k)},\bm{\eta}^{(k)},\bm{Z}^{(k)},\bm{z}_1^{(k)},\bm{z}_2^{(k)}\}$, we update them by alternating optimization. 
When updating the primal variable $\bm{P}$, we can ignore Constraint 3 and the three regularizers (because they are unrelated to $\bm{P}$) and write the Constraint 2 explicitly. 
Then, the problem becomes:
\begin{eqnarray}\label{eq:updateP}
\begin{aligned}
&\sideset{}{_{\bm{P}\in\Pi(\bm{\mu}^{(k)},\cdot)}}\min L_{\bm{P}}\\
&=\sideset{}{_{\bm{P}\in\Pi(\bm{\mu}^{(k)},\cdot)}}\min\langle-\bm{X},\bm{P}\rangle  + \alpha_0 \text{R}(\bm{P},\bm{S}^{(k)})+\langle\bm{Z}^{(k)},\bm{P}-\bm{S}^{(k)}\rangle + \rho\underbrace{\text{Div}(\bm{P},\bm{S}^{(k)})}_{\text{KL}(\bm{P}|\bm{S}^{(k)})}.
\end{aligned}
\end{eqnarray}
When using the entropic regularizer, $\text{R}(\bm{P},\bm{S}^{(k)})=\langle\bm{S}^{(k)},\log\bm{S}^{(k)}-\bm{1}\rangle$ is a constant. 
Applying the first-order optimality condition, we have
\begin{eqnarray}\label{eq:deriveP}
\begin{aligned}
&\frac{\partial L_{\bm{P}}}{\partial\bm{P}}=\bm{0} \\
&\Rightarrow~~ \rho\log\bm{P}-\bm{X}+\bm{Z}^{(k)}-\rho\log\bm{S}^{(k)}=\bm{0}\\
&\Rightarrow~~ \bm{P}=\exp\left(\frac{\bm{X}-\bm{Z}^{(k)}+\rho\log\bm{S}^{(k)}}{\rho}\right)\\
&\xRightarrow{\text{Project to $\Pi(\bm{\mu}^{(k)},\cdot)$}} \bm{P}^{(k+1)}=\text{diag}(\bm{\mu}^{(k)})\sigma_{\text{row}}\left(\frac{\bm{X}-\bm{Z}^{(k)}+\rho\log\bm{S}^{(k)}}{\rho}\right)\\
&\xRightarrow{\text{Logarithmic Update}}
\log\bm{P}^{(k+1)}=(\log\bm{\mu}^{(k)}-\text{LogSumExp}_{\text{col}}(\bm{Y}))\bm{1}_N^T+\bm{Y},\\
&\text{where}~\bm{Y}=\frac{\rho\log\bm{S}^{(k)}+\bm{X}-\bm{Z}^{(k)}}{\rho}.
\end{aligned}
\end{eqnarray}
When using the quadratic regularizer, we have $\text{R}(\bm{P},\bm{S}^{(k)})=\langle\bm{P},\bm{S}^{(k)}\rangle$. 
We obtain the closed-form solution of $\bm{P}^{(k+1)}$ by a similar way, just computing $\bm{Y}=\frac{\rho\log\bm{S}^{(k)}+\bm{X}-\alpha_0 \bm{S}^{(k)}-\bm{Z}^{(k)}}{\rho}$.

Similarly, when updating the auxiliary variable $\bm{S}$, we ignore the OT Problem, Constraint 2, and Regularizers 2 and 3 and write the Constraint 3 explicitly.
Then, the problem becomes
\begin{eqnarray}\label{eq:updateS}
\begin{aligned}
&\sideset{}{_{\bm{S}\in\Pi(\cdot,\bm{\eta}^{(k)})}}\min L_{\bm{S}}\\
&=\sideset{}{_{\bm{S}\in\Pi(\cdot,\bm{\eta}^{(k)})}}\min \alpha_0\text{R}(\bm{P}^{(k+1)},\bm{S}) +
\langle\bm{Z}^{(k)},\bm{P}^{(k+1)}-\bm{S}\rangle + \rho\underbrace{\text{Div}(\bm{S},\bm{P}^{(k+1)})}_{\text{KL}(\bm{S}|\bm{P}^{(k+1)})}.
\end{aligned}
\end{eqnarray}
When using the entropic regularizer, $\text{R}(\bm{P}^{(k+1)},\bm{S})=\langle\bm{S},\log\bm{S}-\bm{1}\rangle$. 
Applying the first-order optimality condition, we have
\begin{eqnarray}\label{eq:deriveS}
\begin{aligned}
&\frac{\partial L_{\bm{S}}}{\partial\bm{S}}=\bm{0} \\
&\Rightarrow~~ (\alpha_0+\rho)\log\bm{S}-\bm{Z}^{(k)}-\rho\log\bm{P}^{(k+1)}=\bm{0}\\
&\Rightarrow~~ \bm{S}=\exp\left(\frac{\bm{Z}^{(k)}+\rho\log\bm{P}^{(k+1)}}{\alpha_0+\rho}\right)\\
&\xRightarrow{\text{Project to $\Pi(\cdot, \bm{\eta}^{(k)})$}} \bm{S}^{(k+1)}=\sigma_{\text{col}}\left(\frac{\bm{Z}^{(k)}+\rho\log\bm{P}^{(k+1)}}{\alpha_0+\rho}\right)\text{diag}(\bm{\eta}^{(k)})\\
&\xRightarrow{\text{Logarithmic Update}}
\log\bm{S}^{(k+1)}=\bm{1}_D(\log\bm{\eta}^{(k)}-\text{LogSumExp}_{\text{row}}(\bm{Y}))^T +\bm{Y},\\
&\text{where}~\bm{Y}=\frac{\bm{Z}^{(k)}+\rho\log \bm{P}^{(k+1)}}{\alpha_0+\rho}. 
\end{aligned}
\end{eqnarray}
Similarly, when $\text{R}(\bm{P}^{(k+1)},\bm{S})=\langle\bm{P}^{(k+1)},\bm{S}\rangle$, we can derive $\bm{S}^{(k+1)}$ by computing $\bm{Y}=\log \bm{P}^{(k+1)} + \frac{\bm{Z}^{(k)}-\alpha_0 \bm{S}^{(k+1)}}{\rho}$.

When updating the auxiliary variable $\bm{\mu}$, we ignore the OT Problem, Regularizers 1 and 3, Constraints 1 and 3. 
Then, the problem becomes
\begin{eqnarray}\label{eq:update_mu}
\begin{aligned}
\sideset{}{_{\bm{\mu}}}\min L_{\bm{\mu}}=\sideset{}{_{\bm{\mu}}}\min \alpha_1\text{KL}(\bm{\mu}|\bm{p}_0) + \langle\bm{z}_1^{(k)},\bm{\mu}-\bm{P}^{(k+1)}\bm{1}_N\rangle + \rho\underbrace{\text{Div}(\bm{\mu},\bm{P}^{(k+1)}\bm{1}_N)}_{\text{KL}(\bm{\mu}|\bm{P}^{(k+1)}\bm{1}_N)},
\end{aligned}
\end{eqnarray}
where $\bm{P}^{(k+1)}\bm{1}_N$ actually equals to $\bm{\mu}^{(k)}$ because of the constraint in~(\ref{eq:updateP}). 
Therefore, we have 
\begin{eqnarray}\label{eq:derive_mu}
\begin{aligned}
\frac{\partial L_{\bm{\mu}}}{\partial\bm{\mu}}=\bm{0} &\Rightarrow \log\bm{\mu} = \frac{\alpha_1\log\bm{p}_0 +\rho\log\bm{\mu}^{(k)}-\bm{z}_1^{(k)}}{\alpha_1+\rho}
\end{aligned}
\end{eqnarray}

Similarly, when updating the auxiliary variable $\bm{\eta}$, we ignore the OT Problem, Regularizers 1 and 2, Constraints 1 and 2. 
Then, the problem becomes
\begin{eqnarray}\label{eq:update_eta}
\begin{aligned}
\sideset{}{_{\bm{\eta}}}\min L_{\bm{\eta}}=\sideset{}{_{\bm{\eta}}}\min \alpha_1\text{KL}(\bm{\eta}|\bm{q}_0) + \langle\bm{z}_2^{(k)},\bm{\eta}-\bm{S}^{(k+1)T}\bm{1}_D\rangle + \rho\underbrace{\text{Div}(\bm{\eta},\bm{S}^{(k+1)T}\bm{1}_D)}_{\text{KL}(\bm{\eta}|\bm{S}^{(k+1)T}\bm{1}_D)},
\end{aligned}
\end{eqnarray}
where $\bm{S}^{(k+1)T}\bm{1}_D$ actually equals to $\bm{\eta}^{(k)}$ because of the constraint in~(\ref{eq:updateS}). 
Therefore, we have
\begin{eqnarray}\label{eq:derive_eta}
\begin{aligned}
\frac{\partial L_{\bm{\eta}}}{\partial\bm{\eta}}=\bm{0} &\Rightarrow \log\bm{\eta} = \frac{\alpha_2\log\bm{q}_0 +\rho\log\bm{\eta}^{(k)}-\bm{z}_2^{(k)}}{\alpha_2+\rho}
\end{aligned}
\end{eqnarray}

Finally, the dual variables are updated based on the general rule of ADMM algorithm, $i.e.$, 
\begin{eqnarray}\label{eq:zs}
\begin{aligned}
    &\bm{Z}^{(t+1)}=\bm{Z}^{(t)} + \rho(\bm{P}^{(t+1)}-\bm{S}^{(t+1)}),\\
    &\bm{z}_1^{(t+1)}=\bm{z}_1^{(t)} + \rho(\bm{\mu}^{(t+1)} -\bm{P}^{(t+1)}\bm{1}_{N}),\\
    &\bm{z}_2^{(t+1)}=\bm{z}_2^{(t)} + \rho(\bm{\eta}^{(t+1)} -(\bm{S}^{(t+1)})^T\bm{1}_{D}),
\end{aligned}
\end{eqnarray}
which is also applied in~\citep{wang2014bregman,ye2017fast,xu2020gromov}.

\section{More Experimental Results and Implementation Details}\label{app:exp}

\subsection{Basic information of datasets and settings for learning backbone models}
For the backbone models used in each learning task, $e.g.$, the AttentionDeepMIL in~\citep{ilse2018attention} for MIL and the GIN~\citep{xu2018powerful} for graph embedding, we determine their hyperparameters (such as epochs, batch size, learning rate, and so on) based on the typical settings used in existing methods, $i.e.$, Attention-based deep MIL\footnote{\url{https://github.com/AMLab-Amsterdam/AttentionDeepMIL}}~\citep{ilse2018attention} and ADGCL\footnote{\url{https://github.com/susheels/adgcl}}~\citep{suresh2021adversarial}. 
For the ADGCL, we connect the GIN with a linear SVM classifier.
For the hyperparameters of the SVM classifier, we use the default settings shown in the code of the authors. 
In summary, Tables~\ref{tab:data1} and~\ref{tab:data2} show the basic information of the datasets and the settings for learning backbone models. 
It should be noted that all the models (associated with different pooling operations) are trained in 5 trials, and each method uses the same random seed in each trial. 

\begin{table}[t]
\centering
\caption{The basic information of the MIL datasets and the hyperparameters for learning}\label{tab:data1}
\resizebox{\columnwidth}{!}{
\begin{tabular}{c|ccccccc|cccc}
    \hline\hline
    \multirow{3}{*}{Dataset}
    &
    \multicolumn{7}{c|}{Statistics of data}
    &
    \multicolumn{4}{c}{Hyperparameters}
    \\
    & 
    Instance  & 
    \#total    & 
    \#positive    & 
    \#negative    & 
    \#total    & 
    Minimum & 
    Maximum &
    \multirow{2}{*}{Epochs} &
    Batch  &
    Learning &
    Weight
    \\
    & 
    dimension  & 
    bags    & 
    bags    & 
    bags    & 
    instances    & 
    bag size & 
    bag size &
    &
    size  &
    rate  &
    decay
    \\
    \hline
    Messidor &
    687 &
    1200 &
    654 &
    546 &
    12352 &
    8 &
    12 &
    50 &
    128 &
    0.0005 &
    0.005
    \\
    Component &
    200 &
    3130 &
    423 &
    2707 &
    36894 &
    1 &
    53 &
    50 &
    128 &
    0.0005 &
    0.005
    \\
    Function &
    200 &
    5242 &
    443 &
    4799 &
    55536 &
    1 &
    51 &
    50 &
    128 &
    0.0005 &
    0.005
    \\
    Process &
    200 &
    11718 &
    757
     &
     10961
     &
    118417 &
    1 &
    57 &
    50 &
    128 &
    0.0005 &
    0.005
    \\
    \hline\hline
    \end{tabular}
}
\end{table}

\begin{table}[t]
\centering
\caption{The basic information of the graph datasets and the hyperparameters of ADGCL}\label{tab:data2}
\resizebox{\columnwidth}{!}{
\begin{threeparttable}
    \begin{tabular}{c|cccc|ccccc}
    \hline\hline
    \multirow{3}{*}{Dataset}   &
    \multicolumn{4}{c|}{Statistics of data} &
    \multicolumn{5}{c}{Hyperparameters of ADGCL}
    \\
    &
    \multirow{2}{*}{\#Graphs} & 
    Average  & 
    Average  & 
    \multirow{2}{*}{\#Classes}  &
    Node attribute &
    Augmentation &
    \multirow{2}{*}{Epochs} &
    Batch &
    Learning
    \\
    &
    & 
    \#nodes  & 
    \#edges  & 
    &
    dimension &
    methods* &
    &
    size &
    rate
    \\
    \hline
    NCI1 &
    4110 &
    29.87 &
    32.30 &
    2 &
    1 &
    LED
    &
    20
    &
    32
    &
    0.001
    \\
    PROTEINS &
    1113 &
    39.06 &
    72.82 &
    2 &
    1 &
    LED
    &
    20
    &
    32
    &
    0.001
    \\
    MUTAG &  
    188 &
    17.93 &
    19.79 &
    2 &
    1 &
    LED
    &
    20
    &
    32
    &
    0.001
    \\
    COLLAB  & 
    5000 &
    74.49 &
    2457.78 &
    3 &
    1 &
    LED
    &
    100
    &
    32
    &
    0.001
    \\
    RDT-B &
    2000 &
    429.63 &
    497.75 &
    2 &
    1 &
    LED
    &
    150
    &
    32
    &
    0.001
    \\
    RDT-M5K &
    4999 &
    508.52 &
    594.87 &
    5 &
    1 &
    LED
    &
    20
    &
    32
    &
    0.001
    \\
    IMDB-B &
    1000 &
    19.77 &
    96.53 &
    2 &
    1 &
    LED
    &
    20 
    &
    32
    &
    0.001
    \\
    IMDB-M &
    1500 &
    13.00 &
    65.94 &
    3 &
    1 &
    LED
    &
    20
    &
    32
    &
    0.001
    \\
    \hline\hline
    \end{tabular}
    \begin{tablenotes}
    \item[*] ``LED'' for learnable edge drop.
    \end{tablenotes}
\end{threeparttable}
}
\end{table}

\subsection{Settings of pooling layers}
For the pooling layers used in our experiments, some of them are parametrized by attention modules, and thus, need to set hidden dimension $h$. 
For these pooling layers, we use their default settings shown in the corresponding references~\citep{ilse2018attention,yan2018deep,lee2016generalizing}. 
Specifically, we set $h=64$ in the MIL experiment and $h=32$ in the graph embedding experiment, respectively.

Additionally, as aforementioned, the configurations of our UOTP layers include $i$) the number of stacked modules $K$; $ii$) fixing or learning $\bm{p}_0$ and $\bm{q}_0$; $iii$) whether predefining $\alpha_0$ for the Sinkhorn-based UOTP for avoiding numerical instability. 
Table~\ref{tab:uotp_setting} lists the configurations used in our experiments. 
We can find that our UOTP layers are robust to their hyperparameters in most situations, which can be configured easily. 
In particular, in most situations, we can simply set $\bm{p}_0$ and $\bm{q}_0$ as fixed uniform distributions, $K=4$ or $8$, and make $\alpha_0$ unconstrained for the BADMM-based UOTP layers. 
\textbf{In the cases that the Sinkhorn-based UOTP is unstable, we have to set $\alpha_0$ as a large number.} 
\begin{table}[t]
\centering
\caption{The configurations of our UOTP layers}\label{tab:uotp_setting}
\begin{small}
\begin{threeparttable}
    \begin{tabular}{c|c|cccc|cccc}
    \hline\hline
    \multirow{2}{*}{Task}   &
    \multirow{2}{*}{Dataset}   &
    \multicolumn{4}{c|}{UOTP$_{\text{Sinkhorn}}$} &
    \multicolumn{4}{c}{UOTP$_{\text{BADMM-E/B}}$} \\
    &
    &
    $\alpha_0$ &
    $\bm{p}_0$ &
    $\bm{q}_0$ &
    $K$ &
    $\alpha_0$ &
    $\bm{p}_0$ &
    $\bm{q}_0$ &
    $K$ \\
    \hline
    \multirow{4}{*}{MIL} &
    Messidor &
    ---
    &
    Fixed
    &
    Fixed
    &
    4
    &
    ---
    &
    Fixed
    &
    Fixed
    &
    4
    \\
    &
    Component &
    ---
    &
    Fixed
    &
    Fixed
    &
    4
    &
    ---
    &
    Fixed
    &
    Fixed
    &
    4
    \\
    &
    Function &
    ---
    &
    Fixed
    &
    Fixed
    &
    4
    &
    ---
    &
    Fixed
    &
    Fixed
    &
    4
    \\
    &
    Process &
    ---
    &
    Fixed
    &
    Fixed
    &
    4
    &
    ---
    &
    Fixed
    &
    Fixed
    &
    4
    \\
    \hline
    &
    NCI1 &
    ---
    &
    Fixed
    &
    Fixed
    &
    4
    &
    ---
    &
    Fixed
    &
    Fixed
    &
    4
    \\
    &
    PROTEINS &
    2000
    &
    Fixed
    &
    Fixed
    &
    4
    &
    ---
    &
    Fixed
    &
    Fixed
    &
    4 
    \\
    &
    MUTAG &
    ---
    &
    Fixed
    &
    Fixed
    &
    4 
    &
    ---
    &
    Fixed
    &
    Fixed
    &
    4
    \\
    
    Graph &
    COLLAB &
    $10^{10}$
    &
    Fixed
    &
    Fixed
    &
    4
    &
    ---
    &
    Fixed
    &
    Fixed
    &
    4
    \\
    Embedding &
    RDT-B &
    $10^{12}$
    &
    Fixed
    &
    Fixed
    &
    4
    &
    ---
    &
    Fixed
    &
    Fixed
    &
    4
    \\
    &
    RDT-M5K &
    $10^{10}$
    &
    Fixed
    &
    Fixed
    &
    4
    &
    ---
    &
    Fixed
    &
    Fixed
    &
    4
    \\
    &
    IMDB-B &
    $10^{12}$
    &
    Fixed
    &
    Fixed
    &
    4
    &
    ---
    &
    Fixed
    &
    Fixed
    &
    4
    \\
    &
    IMDB-M &
     $10^{11}$
    &
    Fixed
    &
    Fixed
    &
    4
    &
    ---
    &
    Fixed
    &
    Fixed
    &
    4
    \\
    \hline\hline
    \end{tabular}
    \begin{tablenotes}
    \item[1] ``---'' means $\alpha_0$ is a learnable parameters.
    \end{tablenotes}
\end{threeparttable}
\end{small}
\end{table}

\subsection{More experimental results}

\begin{figure}[t]
    \centering
    \includegraphics[height=5cm]{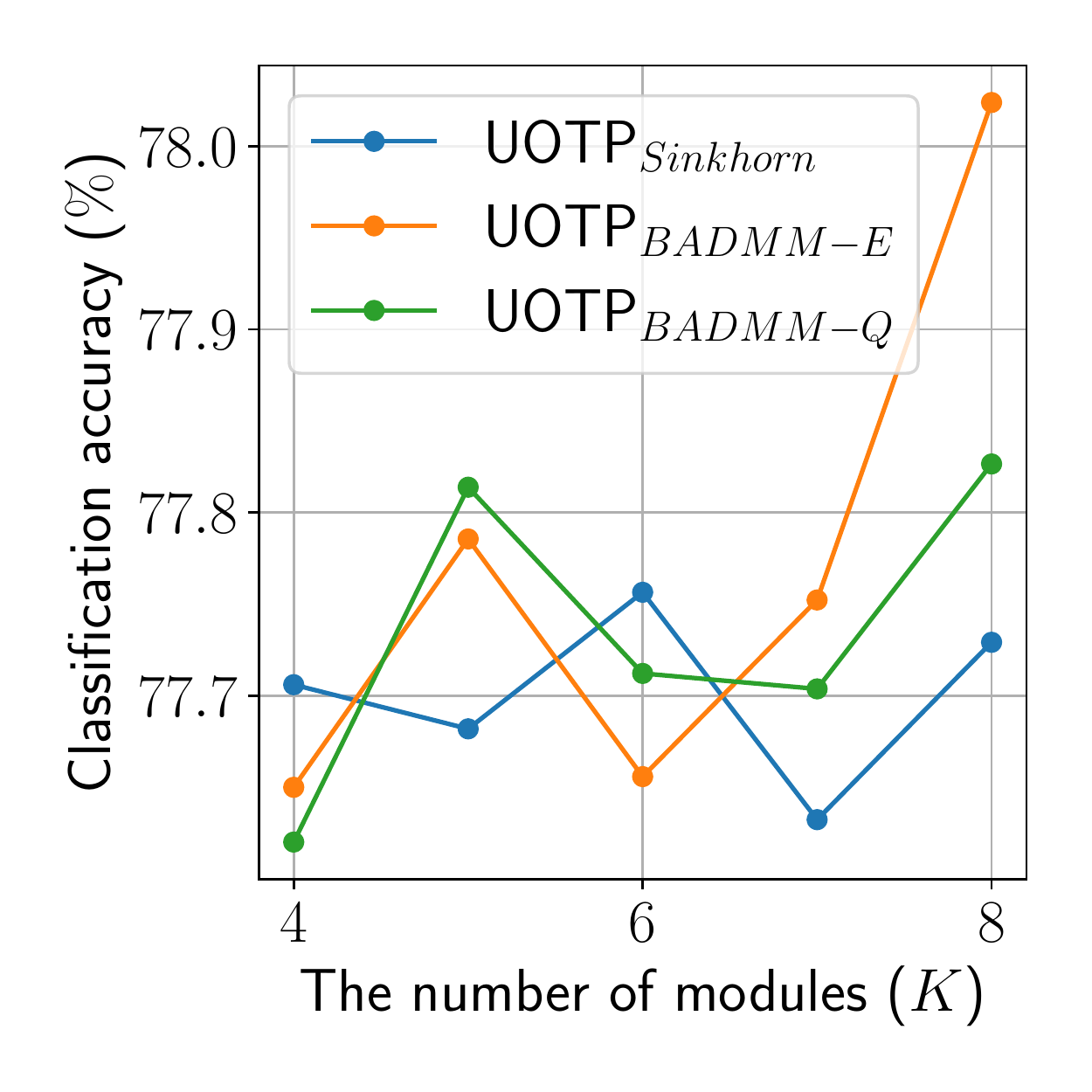}
    \caption{The averaged classification accuracy for the 12 datasets achieved by our UOTP layers under different $K$'s.}
    \label{fig:acc}
\end{figure}

\textbf{Robustness to $K$.} 
Our UOTP layers are simple and robust. 
Essentially, they only have one hyperparameter --- the number of stacked modules $K$. 
Applying a large $K$ will lead to highly-precise solutions to~\eqref{eq:uot} but take more time on both feed-forward computation and backpropagation. 
Fortunately, in most situations, our UOTP layers can obtain encouraging performance with small $K$'s, which achieves a good trade-off between effectiveness and efficiency. 
Figure~\ref{fig:acc} shows the averaged classification accuracy of different UOTP layers on the 12 datasets with respect to $K$'s. 
The performance of our UOTP layers is stable --- when $K\in [4, 8]$, the change of the averaged classification accuracy is smaller than 0.4\%.
This result shows the robustness to the setting of $K$. 

\textbf{Robustness to prior distributions' settings.}
Besides $K$, we also consider the settings of the prior distributions ($i.e.$, $\bm{p}_0$ and $\bm{q}_0$). 
As mentioned in Section~\ref{ssec:implement}, we can fix them as uniform distributions or learn them as parametric models. 
Take the NCI1 dataset as an example. 
Table~\ref{tab:prior} presents the learning results of our methods under different settings of $\bm{p}_0$ and $\bm{q}_0$. 
We can find that our UOT-Pooling layers are robust to their settings --- the learning results do not change a lot under different settings. 
Therefore, in the above experiments, we fix $\bm{p}_0$ and $\bm{q}_0$ as uniform distributions. 
Under this simple setting, our pooling methods have already achieved encouraging results.

\begin{table}[t]
\caption{The impacts of $\bm{p}_0$ and $\bm{q}_0$ on classification accuracy (\%)}\label{tab:prior}
\centering
\begin{small}
    \begin{tabular}{c|ccc|c}
    \hline\hline
    Layer
    & $\bm{p}_0$ 
    & $\bm{q}_0$
    & $K$
    & NCI1
    \\
    \hline
    \multirow{4}{*}{Sinkhorn} 
    & Fixed
    & Fixed 
    & 4
    & 68.27$_{\pm\text{1.06}}$
    \\
    & Learned
    & Fixed
    & 4
    & 67.97$_{\pm\text{0.48}}$
    \\
    & Fixed
    & Learned 
    & 4
    & 69.86$_{\pm\text{0.45}}$
    \\
    & Learned
    & Learned 
    & 4
    & 68.60$_{\pm\text{0.15}}$
    \\
    \hline
    \multirow{4}{*}{BADMM-E}  
    & Fixed
    & Fixed 
    & 4
    & 66.23$_{\pm\text{0.50}}$
    \\
    & Learned
    & Fixed
    & 4
    & 65.96$_{\pm\text{0.22}}$
    \\
    & Fixed
    & Learned
    & 4
    & 66.37$_{\pm\text{0.63}}$
    \\
    & Learned
    & Learned  
    & 4
    & 65.11$_{\pm\text{0.74}}$
    \\
    \hline
    \multirow{4}{*}{BADMM-Q}  
    & Fixed
    & Fixed   
    & 4
    & 66.18$_{\pm\text{0.76}}$
    \\
    & Learned
    & Fixed
    & 4
    & 65.56$_{\pm\text{0.56}}$
    \\
    & Fixed
    & Learned
    & 4
    & 66.24$_{\pm\text{0.89}}$
    \\
    & Learned
    & Learned   
    & 4
    & 65.40$_{\pm\text{0.88}}$
    \\
    \hline\hline
    \end{tabular}
\end{small}
\end{table}

\textbf{The optimal performance achieved by grid search.}
The results in Table~\ref{tab:mil+adgcl} are achieved by setting $K=4$ empirically. 
To explore the optimal performance of our method, for each dataset, we apply the grid search method to find the optimal $K$ in the range $[0, 16]$, and show the results in Table~\ref{tab:opt}. 
We can find that the results of our UOTP layers are further improved.

\begin{table*}[t]
\caption{Comparison on classification accuracy$\pm$Std. (\%) for different pooling layers.}\label{tab:opt}
\resizebox{\textwidth}{!}{
\begin{threeparttable}
    \begin{tabular}{@{}c|c@{\hspace{4pt}}c@{\hspace{4pt}}c@{\hspace{4pt}}c|c@{\hspace{4pt}}c@{\hspace{4pt}}c@{\hspace{4pt}}c@{\hspace{4pt}}c@{\hspace{4pt}}c@{\hspace{4pt}}c@{\hspace{4pt}}c@{}}
    \hline\hline
    \multirow{2}{*}{Pooling}     &
    \multicolumn{4}{c|}{Multi-instance learning} &
    \multicolumn{8}{c}{Graph classification (ADGCL)} 
    \\
    \cline{2-13}
    &
    Messidor &
    Component       & 
    Function       & 
    Process     &
    NCII        & 
    PROTEINS     & 
    MUTAG       & 
    COLLAB       & 
    RDT-B       & 
    RDT-M5K        & 
    IMDB-B    & 
    IMDB-M   \\ 
    \hline
    Add  &  
    74.33$_{\pm\text{2.56}}$
    &
    93.35$_{\pm\text{0.98}}$
    &
    96.26$_{\pm\text{0.48}}$
    & 
    \textbf{97.41$_{\pm\text{0.21}}$}
    &
    67.96$_{\pm\text{0.43}}$
    & 
    72.97$_{\pm\text{0.54}}$
    & 
    \color{red}{\textbf{89.05$_{\pm\text{0.86}}$}}
    &  
    71.06$_{\pm\text{0.43}}$
    & 
    80.00$_{\pm\text{1.49}}$
    &  
    50.16$_{\pm\text{0.97}}$
    &
    70.18$_{\pm\text{0.87}}$  
    & 
    47.56$_{\pm\text{0.56}}$
    \\
    Mean & 
    74.42$_{\pm\text{2.47}}$
    &       
    93.32$_{\pm\text{0.99}}$
    &      
    96.28$_{\pm\text{0.66}}$
    &    
    97.20$_{\pm\text{0.14}}$
    &
    64.82$_{\pm\text{0.52}}$
    &
    66.09$_{\pm\text{0.64}}$
    & 
    86.53$_{\pm\text{1.62}}$
    &  
    72.35$_{\pm\text{0.44}}$
    & 
    83.62$_{\pm\text{1.18}}$
    &  
    \textbf{52.44$_{\pm\text{1.24}}$}
    &
    70.34$_{\pm\text{0.38}}$  
    & 
    48.65$_{\pm\text{0.91}}$ 
    \\
    Max  & 
    73.92$_{\pm\text{3.00}}$
    &   
    93.23$_{\pm\text{0.76}}$
    &       
    95.94$_{\pm\text{0.48}}$
    &      
    96.71$_{\pm\text{0.40}}$
    &
    65.95$_{\pm\text{0.76}}$
    &  
    72.27$_{\pm\text{0.33}}$
    & 
    85.90$_{\pm\text{1.68}}$
    &  
    73.07$_{\pm\text{0.57}}$
    & 
    82.62$_{\pm\text{1.25}}$
    &
    44.34$_{\pm\text{1.93}}$
    &
    70.24$_{\pm\text{0.54}}$
    & 
    47.80$_{\pm\text{0.54}}$  
    \\
    DeepSet  &  
    74.42$_{\pm\text{2.87}}$
    &     
    93.29$_{\pm\text{0.95}}$
    &  
    96.45$_{\pm\text{0.51}}$
    & 
    \color{red}{\textbf{97.64$_{\pm\text{0.18}}$}}
    &
    66.28$_{\pm\text{0.72}}$
    & 
    \color{red}{\textbf{73.76$_{\pm\text{0.47}}$}}
    & 
    87.84$_{\pm\text{0.71}}$
    &  
    69.74$_{\pm\text{0.66}}$
    & 
    82.91$_{\pm\text{1.37}}$
    &  
    47.45$_{\pm\text{0.54}}$
    &
    70.84$_{\pm\text{0.71}}$ 
    & 
    48.05$_{\pm\text{0.71}}$  
    \\
    Mixed  &  
    73.42$_{\pm\text{2.29}}$
    &
    \textbf{93.45$_{\pm\text{0.61}}$}
    &   
    96.41$_{\pm\text{0.53}}$
    &      
    96.96$_{\pm\text{0.25}}$
    &
    66.46$_{\pm\text{0.74}}$
    &  
    72.25$_{\pm\text{0.45}}$
    & 
    87.30$_{\pm\text{0.87}}$
    & 
    \textbf{73.22$_{\pm\text{0.35}}$}
    & 
    \textbf{84.36$_{\pm\text{2.62}}$}
    & 
    46.67$_{\pm\text{1.63}}$
    &
    \textbf{71.28$_{\pm\text{0.26}}$} 
    & 
    48.07$_{\pm\text{0.25}}$  
    \\
    GatedMixed  &
    73.25$_{\pm\text{2.38}}$
    &    
    93.03$_{\pm\text{1.02}}$
    &    
    96.22$_{\pm\text{0.65}}$
    &      
    97.01$_{\pm\text{0.23}}$
    &
    63.86$_{\pm\text{0.76}}$
    &  
    69.40$_{\pm\text{1.93}}$
    & 
    87.94$_{\pm\text{1.28}}$
    &  
    71.94$_{\pm\text{0.40}}$
    & 
    80.60$_{\pm\text{3.89}}$
    &  
    44.78$_{\pm\text{4.53}}$  
    &
    70.96$_{\pm\text{0.60}}$
    & 
    48.09$_{\pm\text{0.44}}$
    \\
    Set2Set  & 
    73.58$_{\pm\text{3.74}}$
    &
    93.19$_{\pm\text{0.95}}$
    &   
    96.43$_{\pm\text{0.56}}$
    &    
    97.16$_{\pm\text{0.25}}$
    &
    65.10$_{\pm\text{1.12}}$
    &  
    68.61$_{\pm\text{1.44}}$
    & 
    87.77$_{\pm\text{0.86}}$
    &  
    72.31$_{\pm\text{0.73}}$
    & 
    80.08$_{\pm\text{5.72}}$
    &  
    49.85$_{\pm\text{2.77}}$
    &
    70.36$_{\pm\text{0.85}}$   
    & 
    48.30$_{\pm\text{0.54}}$
    \\
    Attention   & 
    74.25$_{\pm\text{3.67}}$
    &
    93.22$_{\pm\text{1.02}}$
    &   
    96.31$_{\pm\text{0.66}}$
    &    
    \textbf{97.24$_{\pm\text{0.16}}$}
    &
    64.35$_{\pm\text{0.61}}$
    & 
    67.70$_{\pm\text{0.95}}$
    & 
    88.08$_{\pm\text{1.22}}$
    &  
    72.57$_{\pm\text{0.41}}$
    & 
    81.55$_{\pm\text{4.39}}$
    &  
    51.85$_{\pm\text{0.66}}$
    &
    70.60$_{\pm\text{0.38}}$ 
    & 
    47.83$_{\pm\text{0.78}}$ 
    \\
    GatedAtt  &
    73.67$_{\pm\text{2.23}}$
    &
    \textbf{93.42$_{\pm\text{0.91}}$}
    &   
    \textbf{96.51$_{\pm\text{0.77}}$}
    &    
    97.18$_{\pm\text{0.14}}$
    &
    64.66$_{\pm\text{0.52}}$
    &  
    68.16$_{\pm\text{0.90}}$
    & 
    86.91$_{\pm\text{1.79}}$
    &  
    72.31$_{\pm\text{0.37}}$
    & 
    82.55$_{\pm\text{1.96}}$
    & 
    51.47$_{\pm\text{0.82}}$
    &
    70.52$_{\pm\text{0.31}}$  
    & 
    48.67$_{\pm\text{0.35}}$
    \\
    DynamicP      
    & 
    73.16$_{\pm\text{2.12}}$
    &
    93.26$_{\pm\text{1.30}}$
    &   
    \textbf{96.47$_{\pm\text{0.58}}$}
    &    
    97.03$_{\pm\text{0.14}}$
    &
    62.11$_{\pm\text{0.27}}$
    &  
    65.86$_{\pm\text{0.85}}$
    & 
    85.40$_{\pm\text{2.81}}$
    &  
    70.78$_{\pm\text{0.88}}$
    & 
    67.51$_{\pm\text{1.82}}$
    & 
    32.11$_{\pm\text{3.85}}$
    &
    69.84$_{\pm\text{0.73}}$ 
    &  
    47.59$_{\pm\text{0.48}}$
    \\
    GNP &
    73.54$_{\pm\text{3.68}}$
    &
    92.86$_{\pm\text{1.96}}$
    &   
    96.10$_{\pm\text{1.03}}$
    &    
    96.03$_{\pm\text{0.67}}$
    &
    \textbf{68.20$_{\pm\text{0.48}}$}
    &
    \textbf{73.44$_{\pm\text{0.61}}$}
    &
    88.37$_{\pm\text{1.25}}$
    &
    72.80$_{\pm\text{0.58}}$
    &
    81.93$_{\pm\text{2.23}}$
    &
    51.80$_{\pm\text{0.61}}$
    &
    70.34$_{\pm\text{0.83}}$
    &
    \textbf{48.85$_{\pm\text{0.81}}$}
    \\
    ASAP &
    ---
    &
    ---
    &
    ---
    &
    ---
    &
    \textbf{68.09$_{\pm\text{0.42}}$}
    &
    70.42$_{\pm\text{1.45}}$
    &
    87.68$_{\pm\text{1.42}}$
    &
    68.20$_{\pm\text{2.37}}$
    &
    73.91$_{\pm\text{1.50}}$
    &
    44.58$_{\pm\text{0.44}}$
    &
    68.33$_{\pm\text{2.50}}$
    &
    43.92$_{\pm\text{1.13}}$
    \\
    SAGP &
    ---
    &
    ---
    &
    ---
    &
    ---
    &
    67.48$_{\pm\text{0.65}}$
    &
    72.63$_{\pm\text{0.44}}$
    &
    87.88$_{\pm\text{2.22}}$
    &
    70.19$_{\pm\text{0.55}}$
    &
    74.12$_{\pm\text{2.86}}$
    &
    46.00$_{\pm\text{1.74}}$
    &
    70.34$_{\pm\text{0.74}}$
    &
    47.04$_{\pm\text{1.22}}$
    \\
    \hline
    \multirow{2}{*}{UOTP$_{\text{Sinkhorn}}$} &
    ($K=16$)
    &
    ($K=10$)
    &
    ($K=4$)
    &
    ($K=10$)
    &
    ($K=8$)
    &
    ($K=8$)
    &
    ($K=4$)
    &
    ($K=4$)
    &
    ($K=4$)
    &
    ($K=8$)
    &
    ($K=4$)
    &
    ($K=8$)
    \\
    &
    \color{red}{\textbf{75.92$_{\pm\text{2.39}}$}}
    &
    \color{red}{\textbf{93.67$_{\pm\text{0.80}}$}}
    &   
    \color{red}{\textbf{96.62$_{\pm\text{0.48}}$}}
    &    
    97.18$_{\pm\text{0.15}}$
    &
    \color{red}{\textbf{68.44$_{\pm\text{0.50}}$}}
    &  
    \textbf{73.36$_{\pm\text{0.71}}$}
    & 
    \textbf{88.84$_{\pm\text{1.21}}$}
    &  
    71.20$_{\pm\text{0.55}}$
    & 
    81.54$_{\pm\text{1.38}}$
    &
    52.04$_{\pm\text{1.06}}$
    &
    70.74$_{\pm\text{0.80}}$
    &  
    47.95$_{\pm\text{0.52}}$
    \\
    \multirow{2}{*}{UOTP$_{\text{BADMM-E}}$}  
    &
    ($K=14$)
    &
    ($K=13$)
    &
    ($K=16$)
    &
    ($K=4$)
    &
    ($K=8$)
    &
    ($K=8$)
    &
    ($K=7$)
    &
    ($K=4$)
    &
    ($K=4$)
    &
    ($K=8$)
    &
    ($K=8$)
    &
    ($K=4$)
    \\
    &
    \textbf{75.75$_{\pm\text{2.00}}$}
    &
    93.39$_{\pm\text{0.72}}$
    &   
    96.45$_{\pm\text{0.52}}$
    &    
    97.15$_{\pm\text{0.16}}$
    &
    66.41$_{\pm\text{0.73}}$
    &  
    70.55$_{\pm\text{1.06}}$
    & 
    \textbf{88.95$_{\pm\text{1.01}}$}
    &
    \textbf{73.86$_{\pm\text{0.44}}$}
    & 
    \textbf{86.80$_{\pm\text{1.19}}$}
    &
    \color{red}{\textbf{52.81$_{\pm\text{0.79}}$}}
    &
    \color{red}{\textbf{72.56$_{\pm\text{0.51}}$}}
    &  
    \color{red}{\textbf{ 50.48$_{\pm\text{0.14}}$}}
    \\
    \multirow{2}{*}{UOTP$_{\text{BADMM-Q}}$}
    &
    ($K=11$)
    &
    ($K=16$)
    &
    ($K=8$)
    &
    ($K=4$)
    &
    ($K=4$)
    &
    ($K=14$)
    &
    ($K=5$)
    &
    ($K=4$)
    &
    ($K=8$)
    &
    ($K=4$)
    &
    ($K=4$)
    &
    ($K=8$)
    \\
    &
    \textbf{75.50$_{\pm\text{2.29}}$}
    &
    93.35$_{\pm\text{0.83}}$
    &   
    96.34$_{\pm\text{0.56}}$
    &    
    97.08$_{\pm\text{0.17}}$
    &
    66.18$_{\pm\text{0.76}}$
    &  
    71.77$_{\pm\text{0.85}}$
    & 
    87.92$_{\pm\text{1.11}}$
    &  
    \color{red}{\textbf{74.14$_{\pm\text{0.24}}$}}  
    & 
    \color{red}{\textbf{88.81$_{\pm\text{0.79}}$}}
    &
    \textbf{52.79$_{\pm\text{0.60}}$}
    &
    \textbf{72.34$_{\pm\text{0.50}}$}
    &  
    \textbf{49.81$_{\pm\text{0.64}}$}
    \\ 
    \hline\hline
    \end{tabular}
    \begin{tablenotes}
    \item[*] The top-3 results of each data are bolded and the best result is in red.
    \end{tablenotes}
\end{threeparttable}
}
\end{table*}


\end{document}